\documentclass[pdflatex,sn-mathphys-num]{sn-jnl}% Math and Physical Sciences Numbered Reference Style 
\usepackage{graphicx}%
\usepackage{multirow}%
\usepackage{amsmath,amssymb,amsfonts}%
\usepackage{amsthm}%
\usepackage{mathrsfs}%
\usepackage[title]{appendix}%
\usepackage{xcolor}%
\usepackage{textcomp}%
\usepackage{manyfoot}%
\usepackage{booktabs}%
\usepackage{algorithm}%
\usepackage{algorithmicx}%
\usepackage{algpseudocode}%
\usepackage{listings}%
\usepackage{hyperref,url}
\usepackage{geometry}
\usepackage{longtable}
\usepackage{array}

\theoremstyle{thmstyleone}%
\theoremstyle{thmstyletwo}%

\theoremstyle{thmstylethree}%

\begin{document}

\title{A Proactive Multi-Agent Dialogue Framework for Assessing Social Language Disorder Traits in Autism}
%\title[Article Title]{Proactive Strategy Selection Enhances Diagnostic Efficiency in Autism Language Assessment: A Multi-Agent Framework}
% Proactive Multi-Agent Dialogue Agents for Enhancing the Elicitation of Latent Social Language Disorder Traits in Autism


\author[1]{\fnm{Chuanbo} \sur{Hu}}
\author[1]{\fnm{Minglei} \sur{Yin}}
\author[2]{\fnm{Bin} \sur{Liu}}
\author[1]{\fnm{Wenqi} \sur{Li}}
\author[4]{\fnm{Lynn K.} \sur{Paul}}
\author[3*]{\fnm{Shuo} \sur{Wang}}
\author[1*]{\fnm{Xin} \sur{Li}}\email{xli48@albany.edu}

\affil[1]{\orgdiv{Department of Computer Science}, \orgname{University at Albany}, \city{Albany}, \postcode{12222}, \state{NY}, \country{USA}}

\affil[2]{\orgdiv{Department of Management Information System}, \orgname{West Virginia University},  \city{Morgantown}, \postcode{26506}, \state{WV}, \country{USA}}

\affil[3]{\orgdiv{Department of Radiology}, \orgname{Washington University in St. Louis},  \city{St. Louis}, \postcode{63110}, \state{MO}, \country{USA}}

\affil[4]{\orgdiv{Humanities and Social Sciences}, \orgname{California Institute of Technology}, \city{Pasadena}, \postcode{91125}, \state{CA}, \country{USA}}



\abstract{
%Social Language Disorder (SLD) traits in autism spectrum disorder, characteristic linguistic behaviours 
Characteristic linguistic behaviors associated with Social Language Disorder (SLD) in autism spectrum disorder, 
including echoic repetition, pronoun displacement, and stereotyped media quoting, are largely absent from spontaneous conversation and only emerge under specific conversational conditions. In structured clinical assessments, this latency means that questioning strategy selection is a critical yet underappreciated determinant of how much diagnostic information a conversation yields. Whether large language models (LLMs) can be guided to proactively select questioning strategies that systematically surface these latent traits remains largely unexplored. Here we present TPA (Think, Plan, Ask), a proactive multi-agent dialogue framework applied to the language assessment component of the Autism Diagnostic Observation Schedule Module 4 (ADOS-2), in which a doctor agent explicitly reasons about which traits remain unobserved before selecting a clinically grounded strategy and generating a targeted question. A patient agent grounded in real ADOS-2 clinical data enables reproducible evaluation without real patient participation, validated across three independent experiments confirming adequate fidelity to real patient language. Evaluated on 484 episodes from 35 patients, TPA outperforms six competitive dialogue planning baselines across all primary metrics, achieving 82.1\% SLD trait coverage, 16.6\% points higher than automated replay of real clinical dialogues conducted by trained clinicians (65.5\%), with substantially greater per-turn diagnostic efficiency (AUCC: 0.628 vs. 0.458, absolute gain +0.170). These results demonstrate that proactive questioning strategy selection substantially improves the efficiency of automated SLD trait assessment, with direct implications for scalable AI-assisted clinical screening.}

\maketitle

\section*{Introduction}\label{sec1}

% Long-awaited diagnosis
Autism Spectrum Disorder (ASD) affects approximately 1 in 36 children in the United States \cite{maenner2023}, making it one of the most prevalent neurodevelopmental conditions. Despite decades of research on early detection and intervention, access to timely diagnosis remains deeply inequitable. Across US speciality centres, nearly two-thirds report evaluation wait times exceeding four months, more than one in five have waitlists extending beyond a year or have ceased accepting new referrals, and 77\% of clinics identify assessment length and documentation burden as primary barriers \cite{cognoa2023}. Families from rural areas, low-income backgrounds, and minority communities face even steeper barriers  \cite{doherty2021}. The consequences extend well beyond administrative inconvenience: randomised controlled trial evidence demonstrates that beginning intervention at 18 rather than 27 months produces significantly greater gains in language and adaptive behaviour \cite{guthrie2023}, and systematic reviews confirm that diagnostic delay directly worsens long-term developmental outcomes \cite{daniolou2022}. The average delay from first parental concern to confirmed diagnosis exceeds 27 months in the United States \cite{chen2023}, a gap driven in large part by a shortage of trained evaluators and the time required by the assessments themselves.

% ADOS-2 + prior modality work
The Autism Diagnostic Observation Schedule, Second Edition (ADOS-2)~\cite{lord2012} is the internationally recognised gold standard for ASD assessment. ADOS-2 Module~4, designed for verbally fluent adolescents and adults, requires a trained clinician to conduct a 45--90 minute semi-structured interview, coding social communication behaviours across structured activities. Prior computational work has exploited individual modality of the ADOS-2 interview, including body movement and action recognition from video~\cite{ruan2023mmys}, facial micro-expressions~\cite{ruan2024can}, facial dynamics in interview recordings~\cite{zhang2022discriminative}, social gaze patterns~\cite{yu2024video}, and acoustic speech features~\cite{hu2025speech,fusaroli2017voice}, for automated assessment. Alongside these non-verbal approaches, a parallel line of work has applied natural language processing to ADOS-2 transcripts, extracting automated language measures — including mean length of utterance, word diversity, and repetition proportion — that reliably distinguish ASD from typical development and ADHD across standardised ADOS-2 tasks \cite{salem2021evaluating,macfarlane2022combining}. More recent work has further demonstrated that sentiment and linguistic abstraction features derived from ADOS-2 narrative tasks capture pragmatic impairments with sufficient sensitivity to serve as quantitative complements to clinical coding \cite{chojnicka2020social}. These approaches share a common assumption:  the diagnostic signal of interest can be passively observed from the recording. While this assumption is generally valid for non-verbal modalities, it does not hold for the language component of ADOS-2.

% SLD traits gap
The Social Language Disorder (SLD) traits formally assessed through the A4 algorithm — including echoic repetition, pronoun displacement, stereotyped media quoting, and superfluous phrase attachment — are largely absent from spontaneous conversation and emerge only under specific interactional conditions. This context-dependency is well-established in the ASD language literature: Tager-Flusberg, Paul, and Lord \cite{tager2005language} identified echolalia and pronoun reversal as pragmatic phenomena whose occurrence is tied to conversational register and cognitive load rather than reflecting stable, freely observable traits; Volden and Lord \cite{volden1991neologisms} demonstrated that idiosyncratic language production in verbally fluent autistic speakers increases with utterance complexity and discourse demands, appearing preferentially in naturalistic interview contexts over controlled elicitation tasks; and Luyster et al. \cite{luyster2022conventions} showed that narrative and conversational sampling formats substantially increase the yield of unconventional language forms relative to sentence-level tasks, precisely because such formats impose the social-interactional pressures under which these features surface. Together, these findings establish that reliable elicitation of SLD traits is a prerequisite for valid A4 scoring — a patient who produces no stereotyped language in a poorly structured conversation may receive a score of 0 not because the trait is absent, but because the interaction never created the conditions for it to emerge~\cite{hu2025exploiting}.

%LLM progress + gap
Recent advances in large language models (LLMs) have shown promise across clinical applications, from extracting structured information from clinical records~\cite{stanley2025} and classifying disorder-relevant linguistic features from completed transcripts~\cite{hu2025exploiting}, to conducting general medical interviews with diagnostic quality comparable to primary care physicians~\cite{tu2025towards}. Yet these advances have focused almost entirely on the detection side of the problem --- identifying features from language that has already been produced. The more fundamental challenge of elicitation remains unaddressed: \emph{how to conduct a structured clinical conversation that systematically surfaces latent SLD traits within a constrained number of turns, without the guidance of a trained specialist?} 
%TPA addresses this challenge by combining goal-directed strategy planning with a clinically grounded patient simulator — the two components we show are jointly necessary for efficient elicitation.

\begin{table*}[t]
\centering
\small
\renewcommand{\arraystretch}{1.1}
\caption{Comparison of our proposed TPA (Think, Plan, Ask) framework with representative prior systems (\checkmark\ = supported; \texttimes\ = not supported).}
\label{tab:related_work}
\begin{tabular}{lllllcc}
\toprule
\textbf{Work} & \textbf{Domain} &
\textbf{Patient Sim.} & \textbf{Planning} &
\textbf{Clinical} & \textbf{SLD} \\
\midrule
\textit{Medical Multi-Agent} \\
AI Hospital~\cite{fan2025aihospital}
  & General     & Role-play        & Free-form      & \texttimes & \texttimes \\
AgentClinic~\cite{schmidgall2024agentclinic}
  & Clinical    & Bias-injected    & LLM-Gen        & \texttimes & \texttimes \\
MedAgentSim~\cite{almansoori2025medagentsim}
  & General     & LLM-based        & Experience CoT & \texttimes & \texttimes \\
\midrule
\textit{Psychiatric Simulation} \\
PATIENT-$\Psi$~\cite{wang2024patientpsi}
  & CBT         & Cognitive model  & Human-led      & \checkmark & \texttimes \\
PSYCHE~\cite{kim2025psyche}
  & Psychiatric & Behaviour profile & Free-form     & \texttimes & \texttimes \\
\midrule
\textit{Dialogue Planning} \\
GDP-Zero~\cite{deng2023gdpzero}
  & Persuasion  & N/A              & MCTS           & \texttimes & \texttimes \\
BED-LLM~\cite{choudhury2025bed}
  & Info-seeking & N/A             & Bayesian EIG   & \texttimes & \texttimes \\
DPDP~\cite{dpdp2024}
  & Proactive   & N/A              & Dual-process   & \texttimes & \texttimes \\
UoT~\cite{ye2024uot}
  & Info-seeking & N/A             & Uncertainty TS & \texttimes & \texttimes \\
TOUT~\cite{mo2024tree}
  & Dialogue    & N/A              & Tree-of-Utt.   & \texttimes & \texttimes \\
\midrule
\textbf{TPA (ours)}
  & \textbf{ADOS-2} & \textbf{Trait-grounded}
  & \textbf{Proactive CoT} & \checkmark & \checkmark \\
\bottomrule
\end{tabular}
\end{table*}

As shown in Tab.~\ref{tab:related_work}, prior computational approaches to this problem fall into three broad categories, each addressing a different aspect of the challenge.  \textbf{Medical multi-agent systems} \cite{fan2025aihospital, schmidgall2024agentclinic, almansoori2025medagentsim} have demonstrated the feasibility of LLM-driven clinical dialogue at scale, coordinating multiple agents across general and clinical domains. However, none incorporates trait-level grounding or structured elicitation goals: clinical conversation is treated as a free-form generation task rather than a systematic evidence-gathering process. \textbf{Psychiatric simulation systems} \cite{wang2024patientpsi,kim2025psyche} have moved closer to structured assessment by modelling patient-specific cognitive or behavioural profiles, demonstrating that simulators can reproduce clinically meaningful response patterns. However, these systems rely on human-led or free-form dialogue management without automated strategy selection, and none targets the elicitation of specific linguistic traits as a measurable diagnostic objective. Recent work has applied retrieval-augmented generation to patient simulation~\cite{yu2024aipatient}, grounding LLM responses in real clinical records to improve response fidelity; our Patient Agent extends this principle to strategy-sensitive trait expression, where the simulator must reproduce language conditionally on which questioning strategy is used. Goal-directed dialogue planning methods frame strategy selection as optimisation over conversation states, and have shown promise in persuasion and information-seeking dialogues. However, they share a common limitation: strategy selection is treated as a black-box optimisation problem, disconnected from the clinical logic that determines which conversational conditions will surface which diagnostic features. As a result, none has been applied to structured psychiatric assessment, and none can explain why a particular strategy is chosen at a given moment --- a property that is not only important for clinical transparency, but, as we demonstrate, also drives diagnostic efficiency. Taken together, no existing system combines trait-grounded patient simulation with explicit clinical reasoning over a structured strategy set --- the two components that, as we demonstrate, are jointly necessary for efficient SLD trait elicitation. TPA addresses this gap directly.

\begin{figure}[t]
\centering
\includegraphics[width=\linewidth]{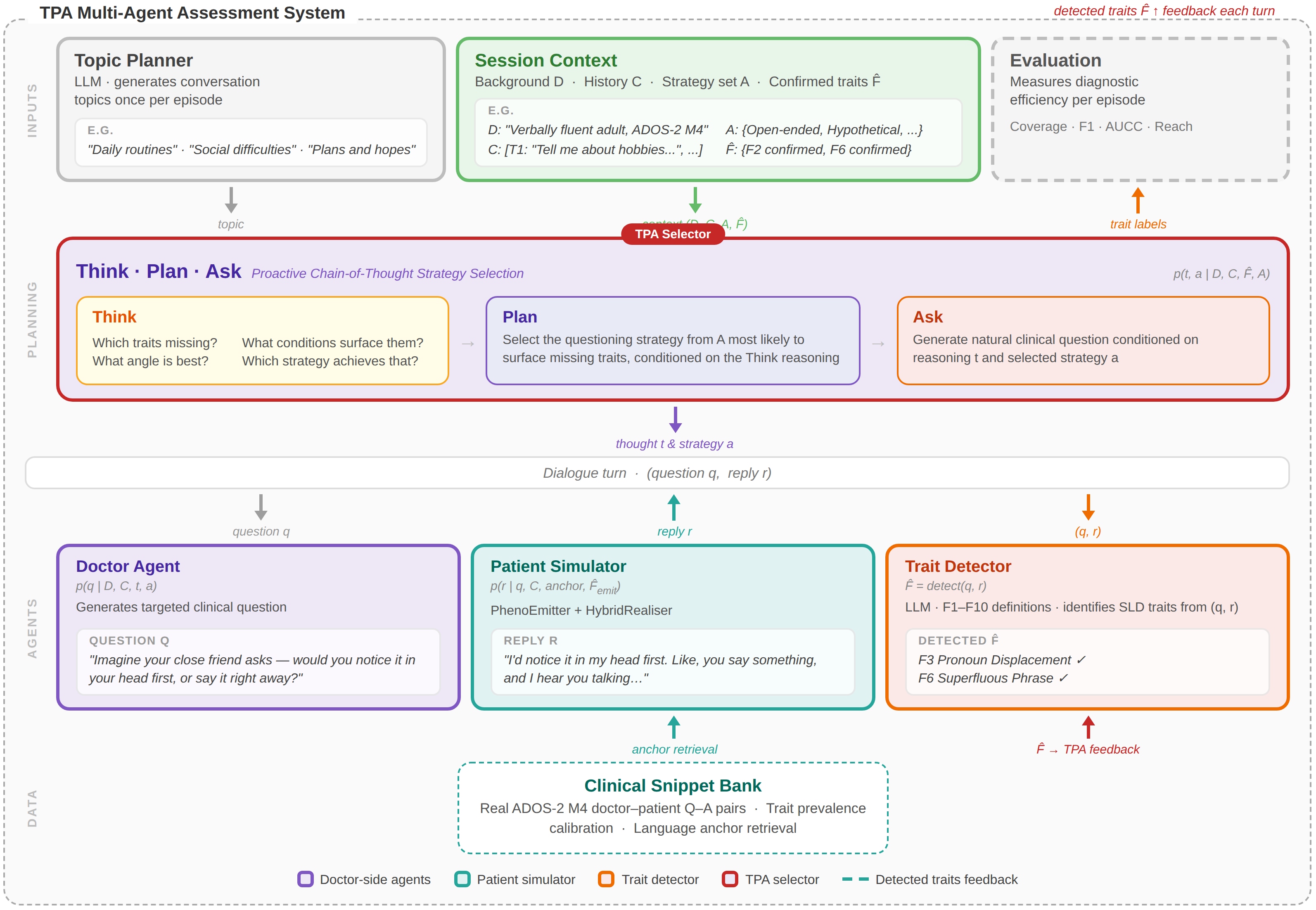}
\caption{Overview of the TPA multi-agent assessment framework, organised across four functional layers. The input layer provides session context to all agents; the planning layer implements the Think-Plan-Ask proactive reasoning cycle; the agent layer comprises the Doctor Agent, Patient Agent, and SLD Trait Detector executing each dialogue turn; and the data layer supplies the Clinical Snippet Bank for trait calibration and language anchor retrieval. Detected traits are fed back to the TPA Selector at each turn, closing the proactive reasoning loop.}
\label{fig:overview}
\end{figure}

%\subsubsection*{TPA System overview}

\begin{figure}[t]
\centering
\includegraphics[width=\linewidth]{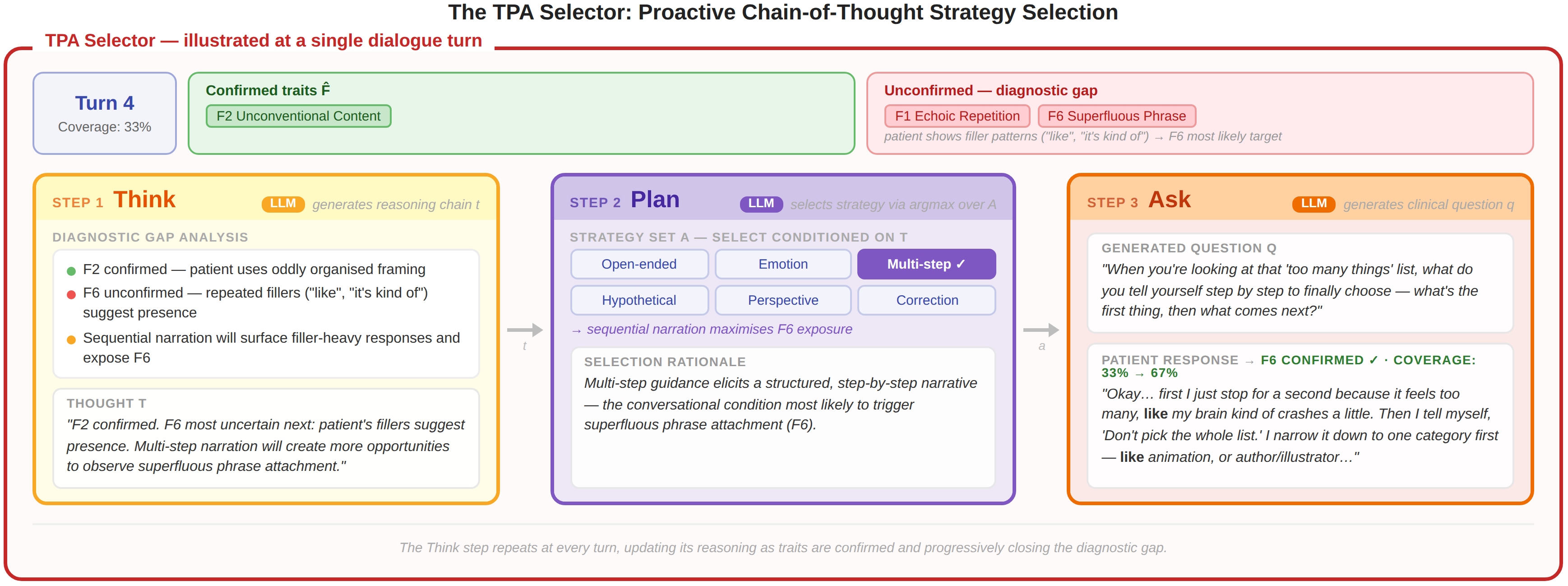}
\caption{The TPA Selector: proactive chain-of-thought strategy selection illustrated at a single dialogue turn. The three steps --- Think, Plan, and Ask --- are executed sequentially at each turn, with each step conditioned on the output of the preceding one. The Think step's reasoning chain drives strategy selection in Plan, which in turn constrains question generation in Ask, ensuring that the clinical question reflects the specific diagnostic intent identified by the reasoning rather than reacting to surface dialogue features alone. The Think step repeats at every turn, progressively closing the diagnostic gap as traits are confirmed.}
\label{fig:TPA_overview}
\end{figure}

\begin{figure}[t]
\centering
\includegraphics[width=\linewidth]{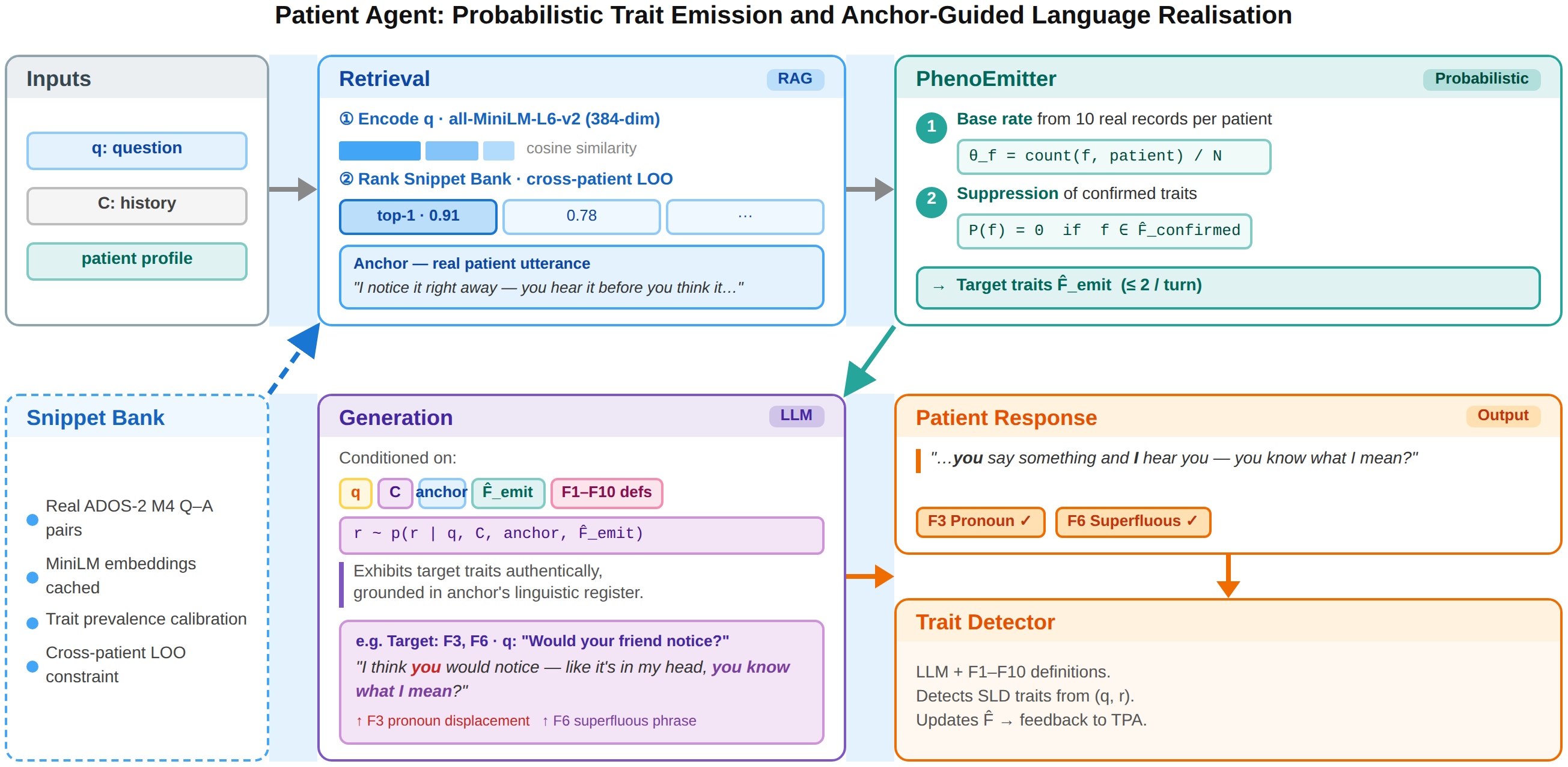}
\caption{Architecture of the Patient Agent, comprising two components: a probabilistic Trait Emission model that determines which SLD traits to express based on patient-level base rates, strategy-specific elicitation profiles, and question–trait semantic affinity; and an LLM-based Language Realiser that generates naturalistic patient responses grounded in real ADOS-2 clinical language via anchor retrieval from the Clinical Snippet Bank.}
\label{fig:patient_simulator}
\end{figure}

Here we present \textbf{TPA} (\textbf{T}hink, \textbf{P}lan, \textbf{A}sk), a proactive multi-agent dialogue framework (Fig. \ref{fig:overview}) designed to improve diagnostic efficiency in structured autism language assessment through systematic questioning strategy selection. The central clinical insight motivating TPA is that expert clinicians do not formulate questions reactively: before each utterance, they reason about what diagnostic evidence is still missing, why it has not yet emerged, and which conversational approach is most likely to surface it. TPA formalises this reasoning cycle as three explicit steps: 1) \emph{Think}, which generates a structured analysis of the current diagnostic state and identifies priority unobserved traits; 2) \emph{Plan}, which selects the optimal questioning strategy from a clinically grounded action set conditioned on this analysis; and 3) \emph{Ask}, which generates a targeted clinical question conditioned on both, thereby transforming reactive conversation into systematic diagnostic elicitation. 

Evaluating such a framework requires a patient model that responds to different questioning strategies in the way real patients do. We therefore develop a patient agent grounded in real ADOS-2 clinical data, combining a probabilistic trait emission model calibrated from clinical transcripts with a hybrid language realiser anchored to authentic patient responses. Three independent validation experiments confirm that the simulator reproduces real clinical language with adequate fidelity across trait frequency, and semantic content. The primary contributions of this work are summarized as follows:

\begin{itemize}
    \item Proactive Multi-Agent Framework: We propose TPA (Fig. \ref{fig:TPA_overview}), a novel proactive multi-agent dialogue framework that formalizes the diagnostic reasoning process into explicit Think-Plan-Ask cycles. By reasoning about unobserved diagnostic gaps, TPA transforms automated assessment from passive observation into systematic, goal-directed elicitation.
    
    \item Clinically-Grounded Patient Agent: We develop and validate a patient agent grounded in real ADOS-2 clinical data (Fig. \ref{fig:patient_simulator}). Through three independent validation experiments, we demonstrate that the simulator faithfully reproduces the linguistic patterns and strategy-dependent behaviors of real patients, providing a reproducible environment for clinical dialogue research.
    
    \item Superior Diagnostic Efficiency: We demonstrate that TPA substantially outperforms six competitive dialogue planning baselines. Crucially, compared to real-world clinical dialogues, TPA achieves an absolute gain of +0.170 in per-turn diagnostic efficiency (AUCC: 0.628 vs. 0.458) and 16.6 percentage points higher trait coverage (82.1\% vs. 65.5\%), significantly compressing the diagnostic window.

\end{itemize}

By formalising the clinical reasoning cycle of expert ADOS-2 assessors into explicit Think, Plan, and Ask steps, TPA transforms automated autism language assessment from passive observation into systematic, goal-directed elicitation. The results demonstrate that proactive questioning strategy selection is not merely a practical improvement but a functionally necessary component of efficient diagnostic dialogue — one that black-box planning approaches, however sophisticated, cannot replicate. Beyond autism assessment, the framework offers a generalisable architecture for any structured clinical interview in which the relevant evidence must be actively created rather than passively observed.

\section*{Results}

To evaluate the efficacy of the TPA framework in eliciting SLD traits, we conducted a series of computational experiments comparing proactive strategy selection against competitive dialogue planning baselines and real-world clinical performance. Our evaluation focuses on whether explicit diagnostic reasoning can improve the diagnostic efficiency, the rate and completeness of trait discovery within a constrained conversational window. In the following sections, we first detail the experimental dataset and simulation setup, followed by a validation of our patient agent. We then present a comparative analysis of elicitation efficiency, demonstrating that TPA achieves superior trait coverage and diagnostic gains over both automated baselines and trained clinicians.

\subsection*{Experimental Setup}

\subsubsection*{Dataset}

The evaluation dataset was derived from the Caltech ADOS Dataset~\cite{zhang2022discriminative}, which consists of audio and video recordings of diagnostic ADOS-2 Module 4 interviews conducted with verbally fluent adolescents and adults. The dataset includes 35 participants aged 16 to 37 years (mean age: 24.32), comprising 26 males and 9 females. A subset of 9 participants underwent two ADOS-2 interviews approximately six months apart, yielding a total of 44 interview recordings. All participants received an ASD diagnosis confirmed through expert clinical evaluation using the ADOS-2 scoring system.

\begin{table}[h]
\centering
\caption{Definitions of the ten Social Language Disorder (SLD) traits assessed through the ADOS-2 Module 4 A4 algorithm}
\begin{tabular}{c|l|p{9cm}} % 删除了首尾的 |
\hline
\textbf{F} & \textbf{Name} & \textbf{Explanation} \\ \hline
\(F_1\) & \begin{tabular}[c]{@{}l@{}}Echoic\\ Repetition\end{tabular} & The individual mimics verbatim what has been said by others, including the examiner, or recites phrases from external sources like advertisements or movie scripts, showing a delayed echo response. \\ \hline
\(F_2\) & \begin{tabular}[c]{@{}l@{}}Unconventional\\ Content\end{tabular} & The speech contains peculiarly chosen content or contextually odd phrasing, such as using 'unfreshness through household' for lack of novelty, 'mideast' instead of 'midwest' for U.S. states, or describing entry into a building as 'through various apertures'. \\ \hline
\(F_3\) & \begin{tabular}[c]{@{}l@{}}Pronoun\\ Displacement\end{tabular} & Incorrectly substitutes personal pronouns, using 'you' in place of 'I', or refers to themselves in the third person, either by pronouns like 'he/she' or by their own name. \\ \hline
\(F_4\) & \begin{tabular}[c]{@{}l@{}}Incongruous \\ Humor Timing\end{tabular} & Incorporates humorous or comedic expressions inappropriately during discussions meant to be serious, showing a misalignment between the content's emotional tone and the context. \\ \hline
\(F_5\) & \begin{tabular}[c]{@{}l@{}}Formalistic\\ Language Use\end{tabular} & Employs an overly formal or archaic language style that seems lifted from written texts, legal documents, or old literature, rather than engaging in conversational speech. Examples include elaborate ways of expressing simple ideas or feelings. \\ \hline
\(F_6\) & \begin{tabular}[c]{@{}l@{}}Superfluous \\Phrase Attachment\end{tabular} & Attaches redundant phrases or filler expressions to their speech without contributing any substantive meaning or context, such as 'you know what I mean' or 'as they say,' indicating a habit rather than intentional emphasis. \\ \hline
\(F_7\) & \begin{tabular}[c]{@{}l@{}}Excessive Social\\ Phrasing\end{tabular} & Utilizes conventional social expressions excessively or inappropriately, responding with phrases like 'oh, thank you' in contexts where it does not fit or preempting social gestures not yet performed by the interlocutor. \\ \hline
\(F_8\) & \begin{tabular}[c]{@{}l@{}}Monotone Social\\ Expression\end{tabular} & Reiterates social phrases with an unchanged, monotonous intonation, indicating a lack of genuine emotional engagement or variability in social interactions. \\ \hline
\(F_9\) & \begin{tabular}[c]{@{}l@{}}Stereotyped Media\\ Quoting\end{tabular} & Quotes lines from commercials, movies, or TV shows in a highly stereotypical manner, employing a canned intonation that mimics the original source closely, suggesting a reliance on external media for verbal expressions. \\ \hline
\(F_{10}\) & \begin{tabular}[c]{@{}l@{}}Clichéd Verbal\\ Substitutions\end{tabular} & Resorts to well-known sayings or clichés in lieu of engaging in direct conversational responses, using phrases like 'circle of life' or 'ready to roll' as stand-ins for more personalized communication. \\ \hline
\end{tabular}
\label{tab:traits}
\end{table}

We followed the same scenario-level segmentation and exclusion criteria as \citet{hu2025exploiting}, retaining 11 dialogue-based scenarios per recording after excluding four non-dialogic activities. SLD trait annotations were assigned to each scenario-level unit by trained clinical annotators following ADOS-2 Module 4 A4 scoring guidelines. Each unit was labelled with the subset of ten SLD traits (F1--F10; Tab.~\ref{tab:traits}) present in the patient's language during that scenario. Units with incomplete or ambiguous annotations were excluded from the evaluation set, resulting in a final dataset of 484 scenarios from 35 patients, with one to ten SLD traits annotated per scene. The full set, including those excluded from evaluation, was retained as the clinical snippet bank for patient agent calibration and anchor retrieval.

\subsubsection*{Model Configuration}

The multi-agent framework was implemented using AutoGen~\cite{wu2024autogen} for agent orchestration, with GPT-5.4-nano serving as the backbone LLM for all agents. Each assessment episode ran for a maximum of $T = 20$ dialogue turns. All performance metrics were computed at the episode level and micro-averaged across 484 scenarios samples from patients ($N = 35$), ensuring that patients do not disproportionately influence the results.

\subsubsection*{Baseline Methods}

We compared TPA against six dialogue planning methods spanning distinct computational paradigms. GDP-Zero~\cite{deng2023gdpzero} frames goal-directed dialogue as a Monte Carlo Tree Search problem over dialogue states. DPDP~\cite{dpdp2024} applies dynamic programming to optimise a dialogue policy. UoT~\cite{ye2024uot} introduces uncertainty-aware tree search with Monte Carlo belief updating. TOUT~\cite{mo2024tree} generates multiple candidate utterance sequences through a tree-of-utterances structure. BED-LLM~\cite{choudhury2025bed} selects actions by maximising Bayesian expected information gain. All baseline methods used the same backbone LLM (GPT-5.4-nano), topic planner, patient agent, and trait detector as TPA; differences in results are attributable solely to the strategy selection mechanism.

In addition, we included two reference conditions. \textbf{Random strategy} uniformly samples from the six-element strategy set $\mathcal{A}$ at each turn without any belief state information, providing a non-planning lower bound. \textbf{Real Dialogue} replays actual ADOS-2 clinical transcripts from the dataset through the trait detector and belief state tracking system, providing an empirical reference representing the diagnostic efficiency achieved by trained clinicians under standard assessment conditions.

\subsubsection*{Evaluation Metrics}

We evaluated elicitation performance using two sets of
complementary metrics: episode-level metrics that capture
overall diagnostic completeness and efficiency, and a
strategy-level metric that quantifies the marginal
contribution of individual questioning strategies.

\textbf{Coverage} measures the fraction of ground-truth SLD traits
successfully elicited by the end of an episode, serving as the
primary measure of diagnostic completeness:
\begin{equation}
    \mathrm{Cov}_T = \frac{|\hat{\mathcal{F}}_T \cap \mathcal{F}^*|}{|\mathcal{F}^*|}
\end{equation}
where $\hat{\mathcal{F}}_T$ is the set of traits detected by turn
$T$ and $\mathcal{F}^*$ is the ground-truth trait set.

\textbf{F1 Score} is the harmonic mean of detection precision
and recall, computed over the set of SLD traits detected
within an episode:
\begin{equation}
    \mathrm{F1} = \frac{2 \cdot P \cdot R}{P + R}, \quad
    P = \frac{|\mathrm{TP}|}{|\mathrm{TP}| + |\mathrm{FP}|}, \quad
    R = \frac{|\mathrm{TP}|}{|\mathrm{TP}| + |\mathrm{FN}|}
\end{equation}
where $\mathrm{TP}$ is the set of ground-truth traits correctly
detected, $\mathrm{FP}$ the set of traits detected but absent
from the ground truth, and $\mathrm{FN}$ the set of ground-truth
traits not detected within the episode.

\textbf{AUCC} (Area Under the Coverage Curve) measures
per-turn diagnostic efficiency by averaging coverage across
all turns of an episode:
\begin{equation}
    \mathrm{AUCC} = \frac{1}{T}\sum_{t=1}^{T}\mathrm{Cov}_t
\end{equation}
AUCC rewards systems that surface traits earlier in the
conversation: two systems with identical final coverage
will differ in AUCC if one reaches that coverage in fewer turns.

\textbf{Coverage Gain Rate} quantifies the marginal
contribution of each questioning strategy to trait discovery,
independently of episode-level performance:
\begin{equation}
    \text{GainRate}(a) =
    \frac{|\{t : \text{strategy}_t = a \wedge
    \text{Cov}_t > \text{Cov}_{t-1}\}|}
    {|\{t : \text{strategy}_t = a\}|}
\end{equation}
A turn is counted as effective if using strategy $a$ led to an increase in cumulative coverage --- that is, at least one new ground-truth trait was detected in that turn. Unlike episode-level metrics, Coverage Gain Rate is computed across all episodes and turns jointly, providing a strategy-specific measure of per-turn elicitation success that is independent of turn order or episode context.

% Simulator validity
\subsection*{Patient agent fidelity to real clinical language}

Evaluating elicitation strategies requires a patient agent that reproduces the language patterns of real ADOS-2 patients with sufficient fidelity --- particularly along the dimensions most critical for elicitation evaluation: accurate trait frequency distributions and semantic coherence with authentic patient language. We validated the simulator across two independent experiments using leave-one-out cross-validation, with pre-specified acceptance thresholds for each metric. Tab.~\ref{tab:simval} summarises the results.

\begin{table}[htbp]
\caption{\textbf{Patient agent validation results.}
Three independent experiments using leave-one-out cross-validation
(95\% CI $= \pm 1.96 \times$ SEM).}
\label{tab:simval}
\small
\setlength{\tabcolsep}{6pt}
\begin{tabular}{llcccc}
\toprule
\textbf{Validation dimension} & \textbf{Metric} &
\textbf{Mean $\pm$ SD} & \textbf{Median} &
\textbf{95\% CI}  \\
\midrule
\multirow{3}{*}{Trait frequency alignment}
  & KL Divergence $\downarrow$
  & $0.920 \pm 0.200$ & $0.910$ & $\pm 0.070$ \\
  & Frequency Error $\downarrow$
  & $0.133 \pm 0.051$ & ---     & $\pm 0.018$  \\
  & AUC $\uparrow$
  & $0.750 \pm 0.100$ & $0.780$ & $\pm 0.036$  \\
\midrule
Semantic similarity
  & Cosine similarity $\uparrow$
  & $0.400 \pm 0.080$ & $0.380$ & $\pm 0.028$  \\
\bottomrule
\end{tabular}
\end{table}

Across both validation dimensions, the simulator meets all pre-specified thresholds. Trait frequency alignment is confirmed by AUC $= 0.750$ (threshold $> 0.70$), KL divergence $= 0.920$ (threshold $< 1.0$), and frequency error $= 0.133$ (threshold $< 0.15$). Semantic coherence with authentic patient language is confirmed by cosine similarity $= 0.400$ (threshold $\geq 0.40$). We discuss each dimension in turn.

We assessed whether the simulator correctly reproduces the patient-specific trait prevalence profile --- that is, whether it assigns higher emission probabilities to traits that are genuinely active in each patient. The AUC for discriminating active from inactive traits was $0.750 \pm 0.100$ (median $0.780$; 95\% CI $\pm 0.036$), meeting the pre-specified threshold of $> 0.70$. KL divergence between simulated and real trait frequency distributions was $0.920 \pm 0.200$ (threshold $< 1.0$), and mean frequency error was $0.133 \pm 0.051$ (threshold $< 0.15$), confirming that the overall shape of the trait distribution is faithfully reproduced at the patient level. Per-trait AUC is reported in Tab.~\ref{tab:perauc}.

\begin{table}[htbp]
\caption{\textbf{Per-trait AUC from trait frequency alignment
validation.} AUC measures the simulator's ability to discriminate
active from inactive traits using simulated emission probabilities
as scores. Threshold: AUC $> 0.60$ indicates adequate
discriminability. Traits with $n \leq 3$ patients are excluded
from the overall AUC calculation due to insufficient negative
examples.}
\label{tab:perauc}
\small
\setlength{\tabcolsep}{8pt}
\begin{tabular}{clcc}
\toprule
\textbf{ID} & \textbf{Trait} &
\textbf{AUC Mean $\pm$ SD} & \textbf{Patients ($n$)} \\
\midrule
F1  & Echoic Repetition             & $0.890 \pm 0.070$ & 8  \\
F2  & Unconventional Content        & $0.710 \pm 0.120$ & 29 \\
F3  & Pronoun Displacement          & $0.780 \pm 0.110$ & 25 \\
F6  & Superfluous Phrase Attachment & $0.660 \pm 0.150$ & 31 \\
F10 & Clichéd Verbal Substitution   & $0.980 \pm 0.040$ & 5  \\
\midrule
F4, F5, F7--F9
    & Rare traits
    & $1.000 \pm 0.000$ & 1--3 \\
\bottomrule
\end{tabular}
\end{table}

\textbf{Semantic similarity.}
We assessed whether simulated patient responses share semantic content with real patient language for the same question context. Mean cosine similarity between sentence embeddings of matched simulated and real responses was $0.400 \pm 0.080$ (median $0.380$; 95\% CI $\pm 0.028$), meeting the threshold of $\geq 0.40$ for meaningful semantic alignment. This confirms that the hybrid language realiser successfully grounds generated responses in the semantic register of authentic clinical transcripts, rather than producing language that is superficially fluent but semantically divergent from real patient expression.

% Elicitation efficiency
\subsection*{TPA improves SLD trait elicitation efficiency}
We first compare TPA directly against real ADOS-2 clinical dialogue---the current standard of practice---to establish the clinical significance of proactive strategy selection (Tab.~\ref{tab:tpa_vs_real}). Real Dialogue replays actual interview transcripts from trained clinicians through the same evaluation conditions as TPA, providing an empirical reference for what expert human assessment achieves within the 20-turn budget.

\begin{table}[htbp]
\caption{\textbf{TPA versus real clinical dialogue.}
Comparison of TPA against automated replay of actual ADOS-2 clinical transcripts (Real Dialogue), evaluated on 484 episodes spanning 35 patients (micro-averaged). Bold indicates improvement over Real Dialogue.}
\label{tab:tpa_vs_real}
\small
\setlength{\tabcolsep}{6pt}
\begin{tabular}{lccc}
\toprule
\textbf{Method} &
\textbf{Cov\,$\uparrow$} & \textbf{F1\,$\uparrow$} &
\textbf{AUCC\,$\uparrow$}  \\
\midrule
Real dialogue
  & 65.5\% & 69.1\% & 0.458 \\
Random strategy 
  & 74.7\% & 80.4\% & 0.580  \\
\midrule
TPA (ours)
  & \textbf{82.1\%} & \textbf{85.9\%}
  & \textbf{0.628}  \\
\bottomrule
\end{tabular}
\end{table}

TPA achieves $82.1\%$ trait coverage, compared to $65.5\%$ for Real Dialogue---a gain of 16.6\% points ($+25\%$ relative). AUCC improves from $0.458$ to $0.628$, reflecting substantially higher diagnostic information gain per turn across the full episode. F1 Score is comparable between TPA ($85.9\%$) and Real Dialogue ($69.1\%$), confirming that TPA's higher coverage is not accompanied by a loss of detection precision.

\begin{figure}[htbp]
\centering
\includegraphics[width=0.75\linewidth]{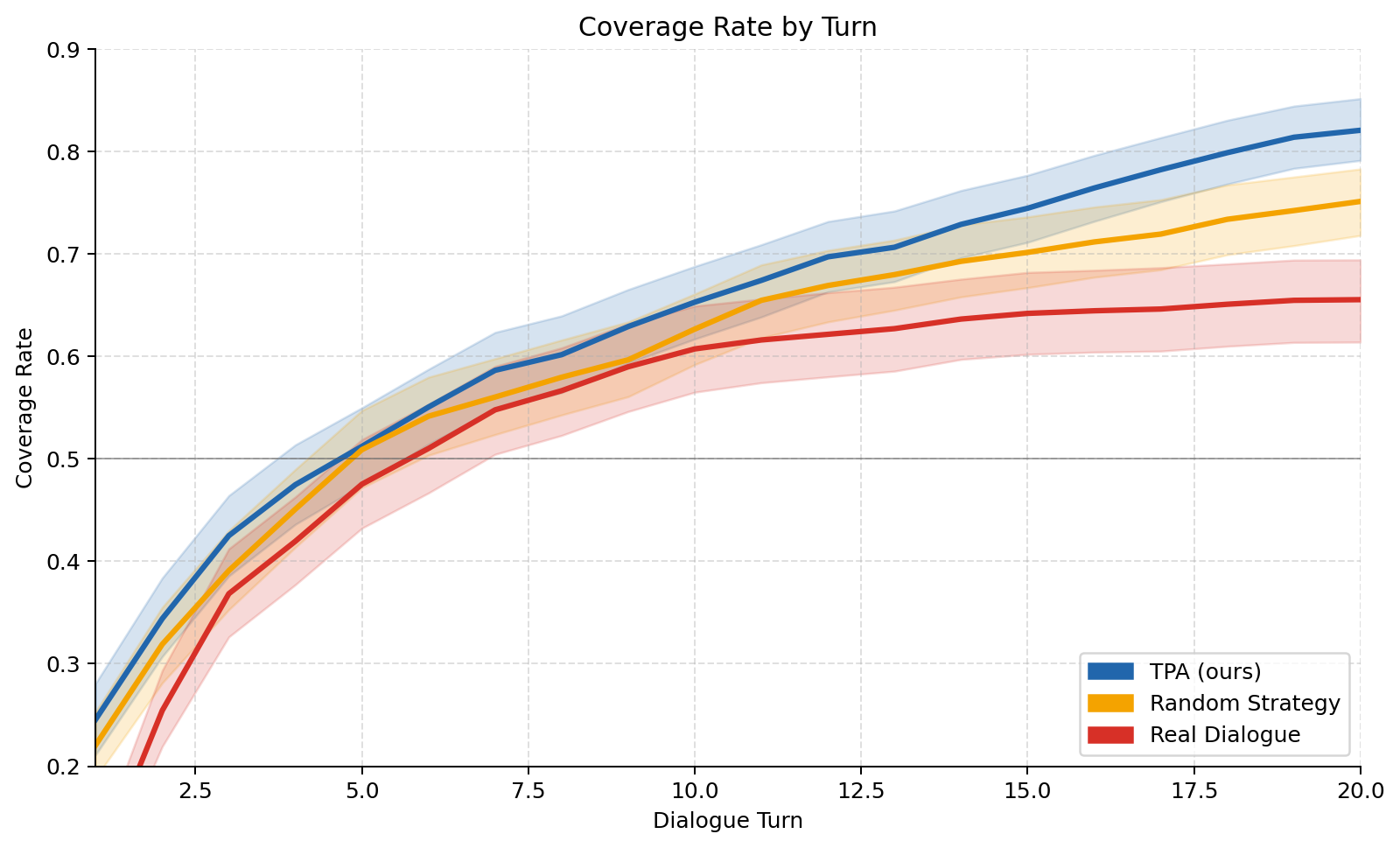}
\caption{\textbf{Coverage trajectory of TPA versus real
clinical dialogue across 20 dialogue turns.} Mean coverage rate ($\pm$ 95\% CI) per turn, micro-averaged over 35 patients. TPA (blue) reaches its inflection point at turn 3 (blue dashed line), compared to turn 11 for real clinical dialogue (red dashed line), reflecting that proactive strategy selection front-loads diagnostic information gain. Thin lines show individual patient trajectories; shaded regions show 95\% CI.}
\label{fig:coverage_curve}
\end{figure}

The coverage trajectory reveals a characteristic pattern of early convergence followed by progressive divergence (Figure~\ref{fig:coverage_curve}). In the first seven turns, all three methods follow closely similar trajectories, reflecting the broad elicitation potential of early-turn questions regardless of strategy. From turn 7 onwards, TPA progressively separates from both baselines: Random strategy plateaus at a moderate rate, while Real Dialogue coverage flattens substantially after turn 10 --- reflecting the time invested in rapport-building, standardised activities, and non-verbal observation that are integral to ADOS-2 but do not directly contribute to SLD trait elicitation. TPA, by contrast, continues to surface new traits in later turns through targeted strategy selection conditioned on the remaining diagnostic gap. This sustained late-turn elicitation is what drives TPA's AUCC advantage: not a faster start, but a more persistent and systematic coverage accumulation across the full episode.

The advantage of TPA is achieved without sacrificing detection precision --- F1 is substantially higher --- suggesting that proactive strategy selection surfaces more traits not by lowering the bar for detection, but by creating the conversational conditions under which traits are more likely to emerge naturally.

% Ablation analysis
\subsection*{Explicit reasoning drives elicitation gains} 

Having established that TPA substantially improves elicitation efficiency over real clinical dialogue, we now examine what drives these gains through two analyses: a comparison against six dialogue planning baselines that isolates the contribution of explicit reasoning over computational search, and a backbone LLM sensitivity analysis that quantifies the dependence of elicitation performance on reasoning quality.

%TPA outperforms all dialogue planning baselines.
Tab.~\ref{tab:main} presents elicitation performance across all methods. TPA achieves the best performance on Coverage ($82.1\%$), weighted F1 ($85.9\%$), and AUCC ($0.628$). The best-performing baseline, TOUT, achieves Coverage of $78.4\%$ and AUCC of $0.605$; TPA surpasses these by 3.7 percentage points and 0.023, respectively, while achieving higher F1 ($85.9\%$ vs.\ $82.8\%$). This margin holds despite TOUT and UoT employing more computationally intensive planning machinery---tree-structured multi-path generation and uncertainty-aware tree search, respectively. Notably, Random strategy ($74.7\%$ Coverage, AUCC $= 0.580$) outperforms several planning baselines on AUCC, suggesting that strategy variation per se provides a baseline benefit that sophisticated but reasoning-free search does not reliably exceed. The further gains of TPA over Random ($+7.4\%$ Coverage, $+0.048$ AUCC) establish that reasoning-guided selection provides a substantial benefit
beyond strategy diversity alone.

\begin{table}[htbp]
\caption{\textbf{Elicitation efficiency across all methods.}
Results micro-averaged at the patient level ($N = 35$,
484 episodes). Bold = best per column. All methods use
GPT-5.4-nano, the same topic planner, patient agent,
and trait detector; differences reflect strategy selection
mechanism only.}
\label{tab:main}
\small
\setlength{\tabcolsep}{4.5pt}
\begin{tabular}{llccc}
\toprule
\textbf{Method} & \textbf{Source} &
\textbf{Cov\,$\uparrow$} & \textbf{F1\,$\uparrow$} &
\textbf{AUCC\,$\uparrow$} \\
\midrule

DPDP\cite{dpdp2024}           & ACL 2024
  & 75.1\% & 80.2\% & 0.611 \\
BED-LLM\cite{choudhury2025bed}      & arXiv 2025
  & 75.6\% & 80.3\% & 0.613  \\
GDP-Zero\cite{deng2023gdpzero} & EMNLP 2023
  & 76.8\% & 81.5\% & 0.611  \\
UoT\cite{ye2024uot}           & NeurIPS 2024
  & 77.9\% & 82.3\% & 0.617  \\
TOUT\cite{mo2024tree}           & ICASSP 2024
  & 78.4\% & 82.8\% & 0.605  \\
\midrule
\textbf{TPA (ours)} & ---
  & \textbf{82.1\%} & \textbf{85.9\%}
  & \textbf{0.628} \\
\bottomrule
\end{tabular}
\end{table}

\begin{figure}[htbp]
\centering
\includegraphics[width=0.75\linewidth]{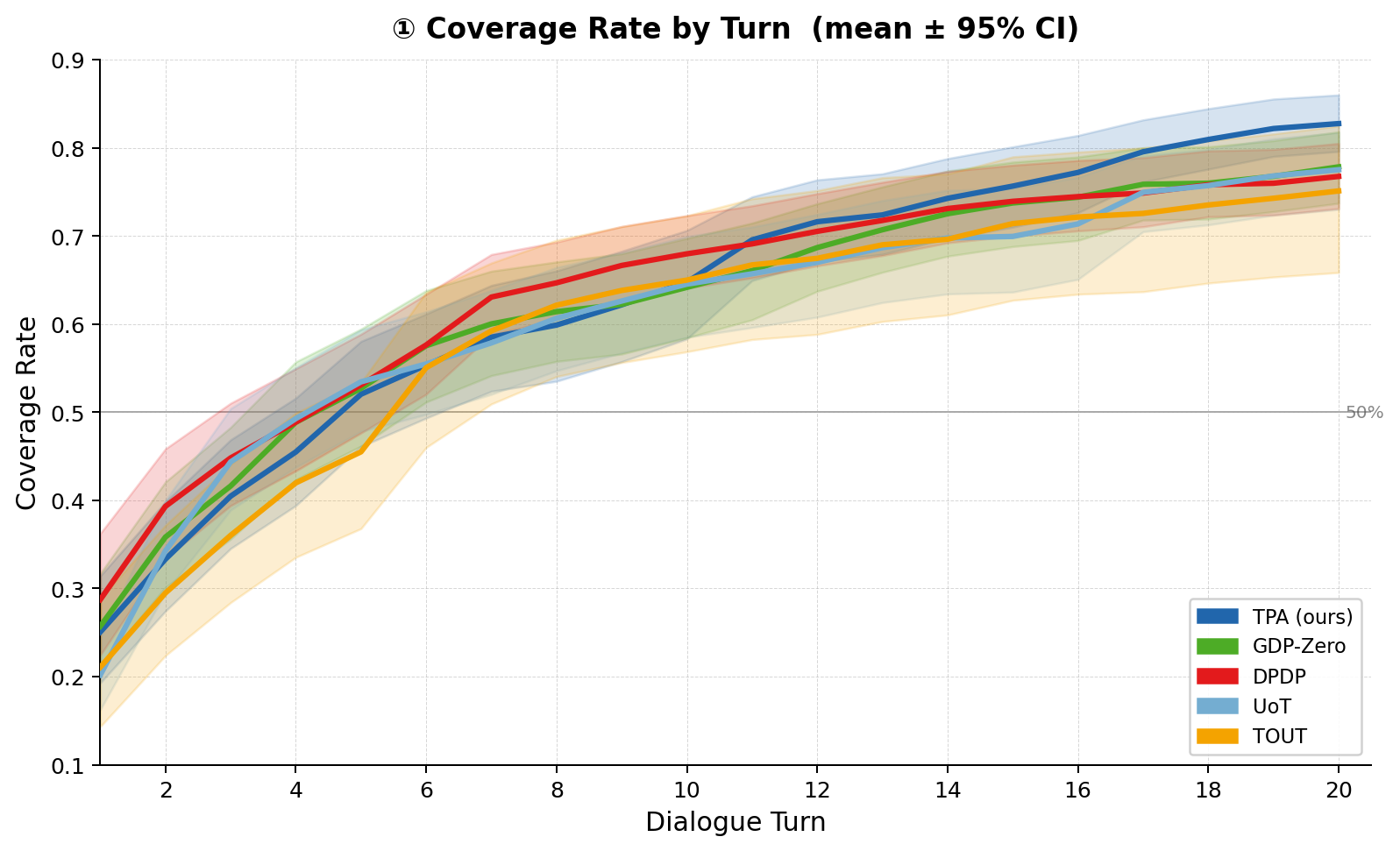}
\caption{\textbf{Coverage trajectories of TPA and five dialogue
planning baselines across 20 dialogue turns.} Mean coverage rate
($\pm$ 95\% CI) micro-averaged across 484 episodes from 35 patients. All methods follow similar trajectories in early turns (1--7), after which
TPA (blue) progressively diverges from the baseline cluster.
TOUT and UoT track TPA most closely among baselines but
consistently fall short. Random strategy outperforms several
planning baselines, confirming that reasoning-guided selection
--- not strategy diversity alone --- drives TPA's advantage.}
\label{fig:coverage_baselines}
\end{figure}

% curve cross all baselines
The coverage trajectories across 20 dialogue turns reveal a consistent advantage of TPA over all baselines throughout the episode (Figure~\ref{fig:coverage_baselines}). TOUT and UoT occupy the upper tier of baselines, tracking TPA most closely but consistently falling short; GDP-Zero and DPDP converge to lower final coverage despite competitive early-turn performance. As noted above, Random strategy outperforms several planning baselines on AUCC, and the further gains of TPA over Random (+7.4\% Coverage, +0.048 AUCC) establish that explicit reasoning provides a substantial benefit beyond strategy diversity alone.

%Reasoning quality scales with model capability.
Tab.~\ref{tab:llm} examines sensitivity to backbone LLM quality, which directly determines the clinical accuracy of the Think step. Coverage scales monotonically from $47.8\%$ (GPT-4o-mini) to $82.1\%$ (GPT-5.4-nano), a gap of 34.3 percentage points. AUCC follows the same pattern ($0.342$ to $0.628$). This steep scaling relationship reflects the clinical reasoning demands of the Think step: a model that cannot accurately identify which traits remain unelicited, or why a specific strategy would surface them, produces Thoughts that misdirect the Plan step regardless of how well Ask executes. The quality of explicit diagnostic reasoning is therefore the hard performance ceiling of the TPA framework.

\begin{table}[htbp]
\caption{\textbf{TPA performance by backbone LLM.} All other components held constant. Performance scales monotonically with model capability, reflecting the clinical reasoning demands of the Think step.}
\label{tab:llm}
\small
\setlength{\tabcolsep}{8pt}
\begin{tabular}{lccc}
\toprule
\textbf{Backbone LLM} &
\textbf{Cov} & \textbf{F1} & \textbf{AUCC} \\

\midrule
GPT-4o-mini         & 47.8\% & 54.1\% & 0.342 \\
GPT-4.1-nano             & 70.6\% & 77.9\% & 0.496 \\
GPT-5.4-nano
  & \textbf{82.1\%} & \textbf{85.9\%} & \textbf{0.628} \\
\bottomrule
\end{tabular}
\end{table}

% Strategy dynamics
\subsection*{Strategy selection is structured and adaptive}

The preceding analyses establish that explicit reasoning about which strategy to use drives elicitation efficiency. We now examine the structure of that strategy space: which strategies contribute most to diagnostic coverage, which traits each strategy preferentially surfaces, and how TPA's strategy selection evolves adaptively across the episode.

\begin{figure}[htbp]
\centering
\includegraphics[width=\linewidth]{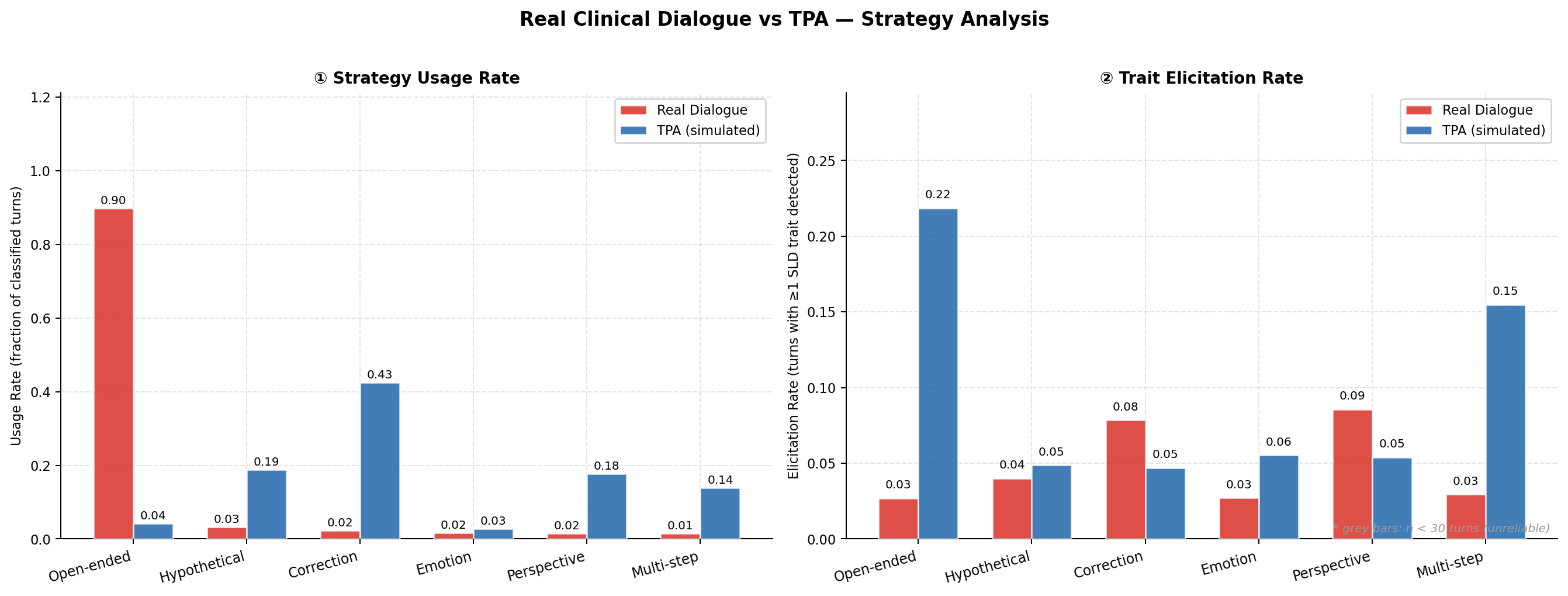}
\caption{Strategy usage and per-turn trait elicitation
rates: real clinical dialogue versus TPA across the first 20 dialogue turns.
\textbf{Left:} Real dialogue concentrates $90\%$ of turns on Open-ended questioning; TPA redistributes turns toward Correction-inducing ($43\%$) and Hypothetical ($19\%$).
\textbf{Right:} Per-turn elicitation rate (fraction of turns with $\geq 1$ SLD trait detected). Open-ended rates are comparable across both conditions ($0.22$ vs $0.03$); strategies with fewer than 30 real dialogue turns are unreliable and shown for reference only. TPA's superior coverage reflects systematic turn reallocation rather than higher per-strategy elicitation rates.}
\label{fig:strategy_comparison}
\end{figure}

We next examine which strategies TPA selects, how their deployment evolves across the episode, and how strategy choice translates into differential trait elicitation. Crucially, this ranking diverges sharply from the strategy distribution observed in real clinical dialogue, where Open-ended questioning dominates $90\%$ of turns, with all remaining strategies collectively accounting for under $10\%$ (Figure~\ref{fig:strategy_comparison}, left). TPA, by contrast, redistributes turns toward Correction-inducing ($43\%$) and Hypothetical ($19\%$) --- strategies deployed in fewer than $3\%$ of real clinical turns each. The per-turn elicitation rates (Figure~\ref{fig:strategy_comparison}, right) reveal that this redistribution is not driven by superior per-strategy performance: Open-ended elicitation rates are substantially higher for TPA than for real dialogue ($0.22$ vs $0.03$), reflecting that TPA deploys Open-ended questions at turns where they are most likely to yield new trait discoveries, rather than uniformly throughout the episode. TPA's superior overall coverage therefore arises not from achieving higher elicitation rates within any individual strategy, but from systematic reallocation of turns toward strategies that target previously unelicited traits at the moments when those traits are most likely to emerge.

\begin{figure}[htbp]
\centering
\includegraphics[width=0.55\linewidth]{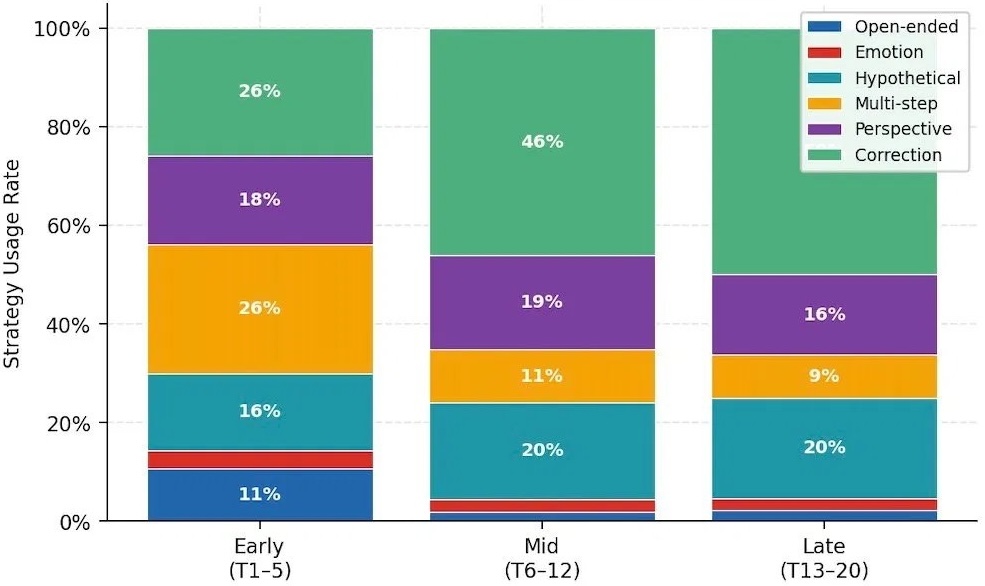}
\caption{\textbf{Strategy distribution across dialogue phases.}
Proportion of turns allocated to each questioning strategy in early (T1--5), mid (T6--12), and late (T13--20) phases, aggregated. Correction-inducing questions rise from $26\%$ in early turns to over $46\%$ in mid and late turns, reflecting the Think step's persistent attempts to elicit echoic repetition (F1) as common traits are confirmed. Multi-step guidance and Open-ended questioning dominate early turns but decline substantially thereafter, consistent with their role in surfacing high-frequency traits (F2, F6) that are typically confirmed within the first five turns.}
\label{fig:strategy_phase}
\end{figure}

To understand when TPA deploys each strategy, Figure~\ref{fig:strategy_phase} decomposes the strategy distribution across three dialogue phases. In early turns (T1--5), TPA distributes effort broadly: Multi-step guidance  ($26\%$) and Hypothetical prompts ($16\%$) dominate alongside Open-ended questioning ($11\%$), reflecting an exploratory phase in which the Think step targets multiple unconfirmed traits simultaneously. From the mid phase (T6--12) onward, Correction-inducing questions rise sharply to $46\%$ and continue to dominate in late turns ($>50\%$), as the diagnostic gap narrows and the Think step redirects attention  toward the most persistently unconfirmed traits --- primarily echoic repetition (F1), which is rare but assigned high belief uncertainty when unobserved. Multi-step guidance and Open-ended questioning decline across phases, consistent with their role in surfacing common traits (F2, F6) that are typically confirmed within the first five turns. This three-phase shift --- from broad exploration to targeted persistence --- is precisely the adaptive pattern that distinguishes TPA from both real clinical dialogue, which concentrates $90\%$ of turns on Open-ended questioning regardless of phase, and from search-based baselines, which do not condition strategy selection on the current diagnostic state.

\subsection*{Case Study}

To illustrate how the Think-Plan-Ask reasoning cycle operates in practice --- and how adaptive strategy selection translates into the coverage gains observed above --- we present a representative episode in which TPA achieves complete trait coverage through three distinct elicitation steps, each driven by an explicit diagnostic reasoning
decision.

\begin{figure}[t]
\centering
\includegraphics[width=1\linewidth]{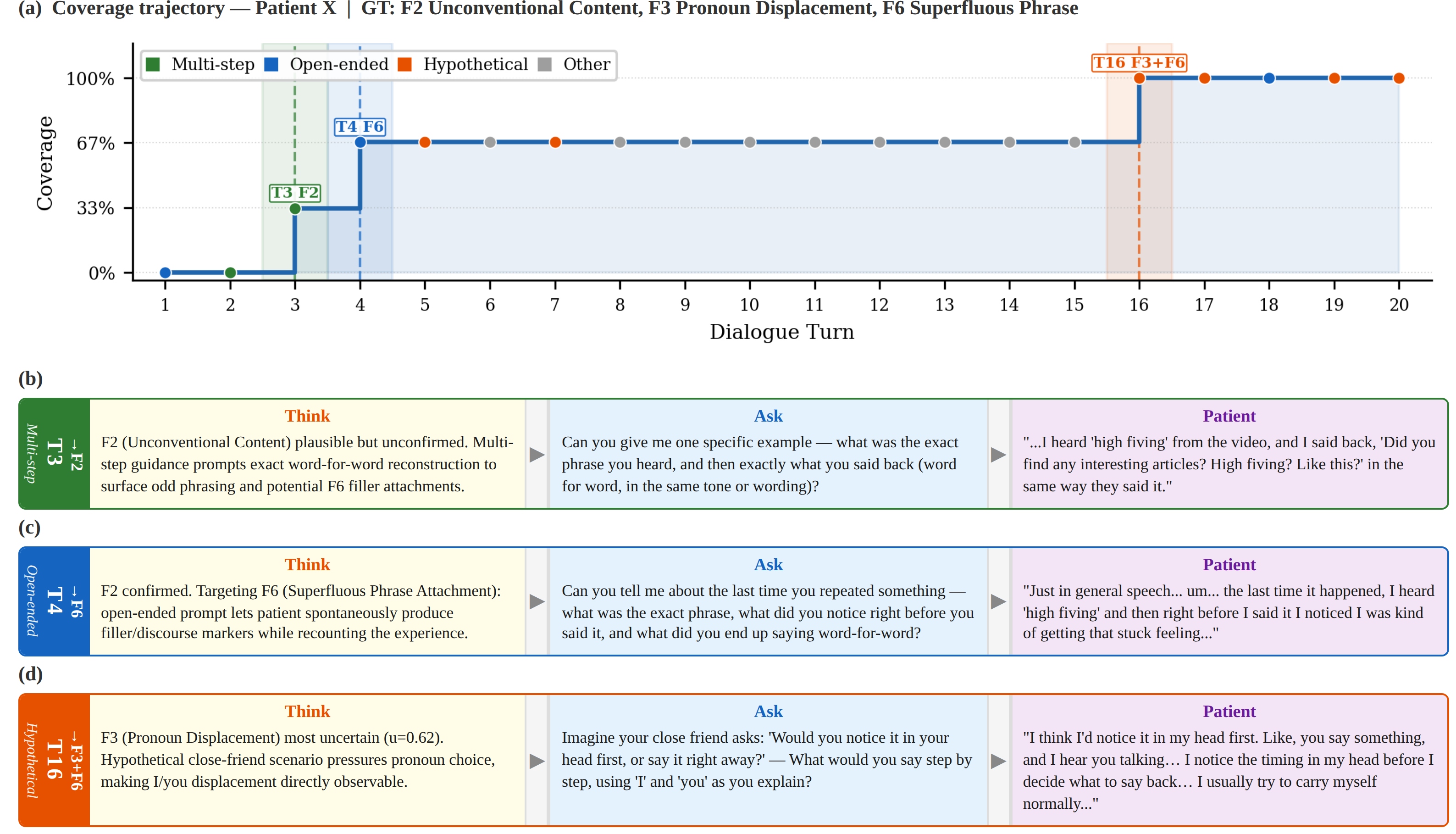}
\caption{Case study: Think-Plan-Ask elicitation cycle for patient
X (GT: F2, F3, F6). (a) Coverage trajectory showing three staircase steps. (b)--(d) Three key turns: Multi-step elicits F2 (T3), Open-ended elicits F6 (T4), Hypothetical surfaces F3+F6 (T16).}
\label{fig:case_study}
\end{figure}

Figure~\ref{fig:case_study} illustrates the Think-Plan-Ask elicitation cycle through a representative episode (Patient~X; GT traits: F2 Unconventional Content, F3 Pronoun Displacement, F6 Superfluous Phrase Attachment; final coverage $= 100\%$, AUCC $= 0.667$). The episode exhibits a characteristic staircase coverage trajectory across three distinct elicitation events, each driven by an explicit diagnostic reasoning step.

At turn~3, the Think step identifies F2 as plausible but unconfirmed and selects Multi-step guidance to prompt word-for-word reconstruction --- creating the conditions under which odd phrasing becomes directly observable. Coverage rises from $0\%$ to $33\%$. At turn~4, with F2 confirmed, the Think step identifies F6 as the next priority and switches to an Open-ended prompt, reasoning that spontaneous retelling will surface filler markers without the structural pressure of a multi-step prompt. Coverage advances to $67\%$. Coverage then plateaus for eleven turns (5--15) as F3 (Pronoun Displacement) remains elusive. At turn~16, the Think step reasons that a hypothetical close-friend scenario will destabilise pronoun use and make I/you displacement directly observable. The Hypothetical prompt surfaces F3 and F6, completing $100\%$ coverage.

This episode illustrates two defining properties of TPA's strategy selection. First, different strategies serve distinct elicitation functions --- Multi-step creates structural conditions for atypical phrasing; Open-ended allows spontaneous filler production; Hypothetical induces pronoun pressure. Second, the Think step adapts across the episode, shifting from broad exploration toward increasingly targeted selection as the diagnostic gap narrows. The resulting staircase trajectory, with each coverage step attributable to a specific strategy-reasoning pair, is precisely the adaptive pattern that underlies TPA's AUCC advantage over both search-based baselines and real clinical dialogue.

%%=============================================================
\section*{Methods}
%%=============================================================

\subsection*{ADOS-2 Module 4}

\begin{table}
\centering
\small
\caption{Overview of scenario tasks in ADOS-2 Module 4.
Scenarios $S_1$, $S_2$, $S_8$, and $S_{10}$ (shaded) are
excluded from the present study as they do not involve
direct examiner--patient dialogue.}
\begin{tabular}{clp{8.5cm}}
\toprule
\textbf{Scenario} & \textbf{Name} & \textbf{Description} \\
\midrule
$S_1$  & Construction Task          & Non-dialogic; tests spatial and motor skills. \\
$S_2$  & Telling a Story from a Book & Non-dialogic; monologic retelling task. \\
$S_3$  & Description of a Picture   & Visual interpretation and description. \\
$S_4$  & Conversation and Reporting & Back-and-forth conversation; past event reporting. \\
$S_5$  & Current Work and School    & Discussion of educational and occupational engagements. \\
$S_6$  & Social Difficulties        & Elicits experiences of social challenges and annoyances. \\
$S_7$  & Emotions                   & Expression and identification of emotions. \\
$S_8$  & Demonstration Task         & Non-dialogic; item demonstration without interaction. \\
$S_9$  & Cartoons                   & Interpretation and explanation of cartoon sequences. \\
$S_{10}$ & Break                    & Non-dialogic; no communicative task. \\
$S_{11}$ & Daily Living             & Discussion of daily routines and personal care. \\
$S_{12}$ & Friends and Relationships & Personal relationships and social norms. \\
$S_{13}$ & Loneliness               & Feelings and experiences of loneliness. \\
$S_{14}$ & Plans and Hopes          & Future aspirations and plans. \\
$S_{15}$ & Creating a Story         & Open-ended creative storytelling. \\
\bottomrule
\end{tabular}
\label{tab:scenarios}
\end{table}

ADOS-2 Module 4 consists of 15 structured activities covering a range of social and communicative contexts (Tab.~\ref{tab:scenarios}). These activities are organised across five assessment categories --- Language and Communication, Reciprocal Social Interaction, Imagination and Creativity, Stereotyped Behaviours and Restricted Interests, and Other Abnormal Behaviours (Tab.~\ref{tab:score}) --- each targeting distinct dimensions of ASD-related behaviour~\cite{lord2012}. The present study focuses exclusively on the A4 item within the Language and Communication category: Stereotyped/Idiosyncratic Use of Words or Phrases, which is scored on a four-point ordinal scale (0--3) reflecting the frequency and pervasiveness of stereotyped or idiosyncratic language (Tab.~\ref{table:A4Score}): from rare or absent (score 0) to predominantly stereotyped speech with little spontaneous flexibility (score 3). Critically, valid A4 scoring requires that the relevant traits have been elicited during the assessment: a patient who produces no stereotyped language receives a score of 0 regardless of whether the absence reflects genuine trait absence or a failure to create the conversational conditions under which the trait would emerge.

\begin{table}[h]
\centering
\small
\caption{Assessment categories of ADOS-2 Module 4. This study focuses on category A, specifically item A4 (Stereotyped/Idiosyncratic Use of Words or Phrases).}
\label{tab:score}
\begin{tabular}{clcp{7.3cm}}
\toprule
\textbf{Class} & \textbf{Name} & \textbf{Items} & \textbf{Description} \\
\midrule
A & Language and Communication      & A1--A10 & Speech and gesture use; clarity, coherence, and appropriateness of language in social interaction. \\
B & Reciprocal Social Interaction   & B1--B13 & Non-verbal and verbal social behaviours, including eye contact, facial expression, and speech quality. \\
C & Imagination / Creativity        & C1      & Use of imagination and creativity in expression and social responses. \\
D & Stereotyped and Restricted Behaviours & D1--D5 & Frequency and intensity of repetitive, restricted, and stereotyped behaviours. \\
E & Other Abnormal Behaviours       & E1--E3  & Overactivity, anxiety, and contextually inconsistent emotional responses. \\
\bottomrule
\end{tabular}
\end{table}

Each of the ten SLD traits (F1--F10; Tab.~\ref{tab:traits}) requires specific conversational conditions to surface reliably. This dependency of each trait on specific elicitation conditions --- and the consequent need to reason about which strategy is most likely to surface each unconfirmed trait at each turn --- is the central clinical motivation for the Think-Plan-Ask design of TPA.

\begin{table}[h]
\centering
\small
\caption{A4 scoring criteria: Stereotyped/Idiosyncratic Use
of Words or Phrases~\cite{lord2012}.}
\label{table:A4Score}
\begin{tabular}{cp{14cm}}
\toprule
\textbf{Score} & \textbf{Description} \\
\midrule
0 & Rarely or never uses stereotyped or idiosyncratic
    language; typical speech patterns throughout. \\
1 & Occasionally uses repetitive or overly formal phrasing,
    though not obviously odd; predominantly spontaneous
    and flexible language. \\
2 & Frequently uses stereotyped or odd words and phrases,
    with some spontaneous language remaining. \\
3 & Predominantly stereotyped or idiosyncratic speech;
    little spontaneous or flexible language use. \\
\bottomrule
\end{tabular}
\end{table}

%The TPA framework
\subsection*{Overview of the multi-agent system framework}

Addressing the elicitation challenge described above requires a system that can reason about what diagnostic evidence is still missing, select the conversational strategy most likely to surface it, and generate a targeted clinical question --- all within the constraints of a naturalistic dialogue. We therefore designed TPA as a multi-agent framework in which these three functions are separated into explicit, inspectable components rather than collapsed into a single end-to-end generation step. This modular separation enables each component to be independently evaluated and replaced, a property important for clinical transparency and iterative system development.

The framework is organised across four functional layers, as illustrated in Figure~\ref{fig:overview}. The \textbf{input layer} provides the session context available to all agents at each turn: clinical background $D$, dialogue history $C$, strategy set $\mathcal{A}$, and the confirmed trait set $\hat{\mathcal{F}}$ accumulated so far. A Topic Planner generates conversation topics once per episode from the ADOS-2 scenario structure, providing scaffolding without exposing ground-truth trait labels. The \textbf{planning layer} contains the TPA Selector --- the core contribution of this work --- which executes an explicit Think-Plan-Ask reasoning cycle at each turn to select a questioning strategy and generate a targeted clinical question. The \textbf{agent layer} comprises three components that execute the dialogue turn: the Doctor Agent generates the clinical question conditioned on the TPA Selector's output; the patient agent generates a naturalistic patient response grounded in real clinical data; and the Trait Detector identifies SLD traits in the patient response, updates $\hat{\mathcal{F}}$, and feeds it back to the TPA Selector, closing the proactive reasoning loop. The \textbf{data layer} consists of the Clinical Snippet Bank --- a repository of real ADOS-2 doctor--patient dialogue pairs --- which the Patient Simulator draws on for trait prevalence calibration and language anchor retrieval. No component has access to ground-truth trait labels during an episode.

The following sections describe the Patient Simulator, TPA Selector, and Trait Detector in the multi-agent system.

\subsection*{Patient Agent: Trait Emission and Language Realisation}

Evaluating elicitation strategies requires a patient model that responds to different questioning strategies in the way real patients do --- one that is sensitive to strategy variation, reproduces authentic clinical language, and correctly reflects the trait profile of each individual patient. We therefore developed a patient agent grounded in real ADOS-2 clinical data, combining an anchor retrieval module with a probabilistic trait emission model and an LLM-based language realiser (Figure~\ref{fig:patient_simulator}).

The agent operates in three sequential stages at each dialogue turn: (1) a \textbf{Retrieval} step that locates the most semantically similar real clinical exchange in the Clinical Snippet Bank to serve as a linguistic anchor; (2) a \textbf{PhenoEmitter} step that determines which SLD traits the response should express, conditioned on the patient profile and current question; and (3) a \textbf{Language Realiser} step that generates a naturalistic patient response grounded in the retrieved anchor and exhibiting the selected traits.

\subsubsection*{Anchor Retrieval}

Before trait emission or response generation, the agent retrieves a real patient utterance from the Clinical Snippet Bank to serve as a linguistic style reference. This retrieval step grounds the generated response in authentic clinical language, preventing the LLM from producing responses that are fluent but stylistically implausible for verbally fluent adults with ASD.

The current doctor question $q$ is encoded using the \texttt{all-MiniLM-L6-v2} sentence encoder~\cite{reimers2019sentence} (384-dimensional embedding), and cosine similarity is computed against all doctor utterances in the Snippet Bank:

\begin{equation}
    \mathrm{anchor} = \arg\max_{s \in \mathcal{B}
    \setminus \mathcal{B}_{\mathrm{patient}}}
    \cos\!\left(
    \mathrm{enc}(q),\; \mathrm{enc}(s.\mathrm{doctor\_curr})
    \right)
    \label{eq:retrieval}
\end{equation}

where $\mathcal{B}$ is the full Snippet Bank and $\mathcal{B}_{\mathrm{patient}}$ denotes the current patient's own records, which are excluded to prevent data leakage under the leave-one-out evaluation design. The paired real patient reply from the top-ranked snippet is used as the anchor response in the subsequent generation step. This cross-patient retrieval constraint ensures that the anchor captures the linguistic register of real ADOS-2 patients without exposing the target patient's own utterances during generation.

\subsubsection*{PhenoEmitter: Trait Emission}

Given the retrieved anchor and the current question $q$, the PhenoEmitter determines which SLD traits the patient response should express. Trait emission is modelled probabilistically, combining two sources of evidence that together determine the emission probability $p_f$ for each trait $f$:

\begin{equation}
    p_f = \sigma\!\left(
        \mathrm{logit}(\theta_f)
        - \mathbb{1}[f \in \hat{\mathcal{F}}] \cdot M
    \right)
    \label{eq:emission}
\end{equation}

The first term, $\theta_f$, is the patient-level base rate --- the empirical frequency with which this patient exhibits trait $f$ in their real ADOS-2 records:

\begin{equation}
    \theta_f = \frac{\mathrm{count}(f,\;\mathrm{patient})}
    {\mathrm{total\_turns}(\mathrm{patient})}
\end{equation}

This term captures individual trait profiles: different patients exhibit different traits at different rates, and the simulator preserves this patient-specificity from the ten real clinical sessions available per patient.

The second term applies a large suppression penalty $M$ when trait $f$ has already been confirmed in the current episode ($f \in \hat{\mathcal{F}}$), preventing repeated emission of already-surfaced traits and ensuring that the simulator does not artificially inflate coverage by re-eliciting known traits. Traits not yet confirmed are emitted according to their base rate $\theta_f$, reflecting the realistic probability that a given question will surface each latent trait.

The target trait set $\hat{\mathcal{F}}_{\mathrm{emit}}$ is sampled from $p_f$ independently for each trait, with a maximum of two traits per turn to reflect the realistic density of SLD trait co-occurrence observed in clinical transcripts.

\subsubsection*{Language Realiser}

Given the retrieved anchor and the target trait set
$\hat{\mathcal{F}}_{\mathrm{emit}}$, the Language Realiser
generates a naturalistic patient response:

\begin{equation}
    r \sim p_{\mathrm{LLM}}\!\left(
        r \mid q,\; C,\; \mathrm{anchor},\;
        \hat{\mathcal{F}}_{\mathrm{emit}},\;
        \Delta_{\mathrm{traits}}
    \right)
    \label{eq:realiser}
\end{equation}

The LLM is conditioned on the current question $q$, dialogue history $C$, the retrieved anchor response, the target trait set $\hat{\mathcal{F}}_{\mathrm{emit}}$, and the full F1--F10 trait definitions $\Delta_{\mathrm{traits}}$. The anchor serves two roles: it provides a lexical and syntactic style reference drawn from authentic clinical speech, and it constrains generation toward the register and vocabulary of real ADOS-2 patients rather than the more formal or generic language an LLM would produce without grounding.

The generated response must satisfy three constraints. First, it must exhibit the target traits naturally and conversationally --- not formulaically insert them as isolated surface markers. Second, it must follow coherently from the prior dialogue history $C$ without abrupt register shifts. Third, it must remain within the linguistic profile established by the anchor, ensuring that responses sound like those of a verbally fluent adult with ASD rather than a generic interlocutor. The generated response $r$ is then passed to the Trait Detector for SLD trait identification, and confirmed traits update the belief state $\hat{\mathcal{F}}$ fed back to the TPA Selector.

The three-stage architecture reflects a deliberate decomposition of the patient simulation problem. The Retrieval stage delegates linguistic style grounding to real clinical data, avoiding the need to fine-tune the LLM on scarce per-patient transcripts --- a task for which the available data (ten sessions per patient) is insufficient for reliable gradient-based alignment. The PhenoEmitter stage separates the when of trait expression (a statistical estimation problem well suited to the available data) from the how (an LLM generation problem that leverages pre-trained linguistic knowledge). 

\subsection*{The TPA Selector: Proactive Chain-of-Thought
Strategy Selection}

The central innovation of TPA is the replacement of reactive question generation with an explicit, turn-by-turn diagnostic reasoning cycle. Rather than generating questions directly from the dialogue history, the TPA Selector first constructs a structured analysis of the current diagnostic state --- identifying which traits remain unconfirmed, why they have not yet emerged, and which conversational conditions would most reliably surface them --- before selecting a strategy and generating a question. The three-step cycle (Think → Plan → Ask, Fig. \ref{fig:TPA_overview}) is executed at every turn, with each step conditioned on the output of the preceding one. At each turn $t$, the TPA Selector receives the full session context: clinical background $D$, dialogue history $C$, strategy set $\mathcal{A}$, and the confirmed trait set $\hat{\mathcal{F}}$ updated by the Trait Detector at the end of the previous turn. 

\subsubsection*{Think: Diagnostic Gap Analysis}

\begin{equation}
    t^* = \arg\max\; \log p(t \mid D,\, C,\, \mathcal{A},\,
    \hat{\mathcal{F}})
    \label{eq:think}
\end{equation}

The Think step generates an explicit reasoning chain $t^*$ that analyses the current diagnostic state before any question is formulated. The LLM is prompted to reason across four dimensions:

\begin{enumerate}
    \item \textbf{Confirmed trait analysis.} Which traits are in $\hat{\mathcal{F}}$, and what do their co-occurrence patterns imply about traits that are likely present but not yet elicited?

    \item \textbf{Diagnostic gap prioritisation.} Which traits in $\mathcal{F} \setminus \hat{\mathcal{F}}$ carry the highest diagnostic uncertainty, as measured by Beta distribution entropy $H(\mathrm{Beta}(\alpha_f, \beta_f))$? The four highest-entropy undetected traits are explicitly injected into the prompt to anchor reasoning and prevent drift toward already-confirmed traits.

    \item \textbf{Elicitation condition analysis.} What conversational conditions --- topic, register, emotional framing, question structure --- are most likely to surface the priority unelicited traits?

    \item \textbf{Strategy identification.} Which strategy from $\mathcal{A}$ would best create those conditions?
\end{enumerate}

The reasoning chain $t^*$ produced by the Think step is not discarded after strategy selection: it is passed directly to the Ask step as a conditioning input, ensuring that the generated question reflects the specific diagnostic intent identified by the reasoning.

\subsubsection*{Plan: Strategy Selection}

\begin{equation}
    a^* = \arg\max_{a \in \mathcal{A}}\; \log p(a \mid D,\,
    C,\, t^*,\, \mathcal{A})
    \label{eq:plan}
\end{equation}

The Plan step selects the questioning strategy $a^*$ from the six-element strategy set $\mathcal{A}$, conditioned on the reasoning chain $t^*$ produced by the Think step. Each strategy in $\mathcal{A}$ encodes a distinct clinical elicitation mechanism, grounded in the trait-strategy dependencies described in the Methods section on ADOS-2 Module 4:

\begin{itemize}
    \item \textbf{Open-ended} --- broad, exploratory prompts suited for early-turn trait profiling before specific unconfirmed traits are identified.
    \item \textbf{Emotion-oriented} --- questions that foreground affective content, preferentially eliciting superfluous phrase attachment (F6) and monotone social
    expression (F8).
    \item \textbf{Hypothetical} --- counterfactual scenario prompts that create self-referential pressure, preferentially eliciting pronoun displacement (F3) and incongruous humour timing (F4).
    \item \textbf{Multi-step} --- sequential, multi-part questions that require elaboration across sub-topics, preferentially eliciting formalistic language (F5) and superfluous phrase attachment (F6).
    \item \textbf{Perspective-taking} --- other-mind reasoning prompts that target social reciprocity failures and theory-of-mind-related language patterns.
    \item \textbf{Correction-inducing} --- questions containing a deliberate mild misstatement, preferentially eliciting echoic repetition (F1) by inviting verbatim correction responses.
\end{itemize}

Strategy selection is conditioned on $t^*$ rather than directly on the session context, ensuring that the choice of strategy reflects the diagnostic reasoning produced by the Think step rather than a pattern-matched response to surface dialogue features.

\subsubsection*{Ask: Targeted Question Generation}

\begin{equation}
    q = p_{\mathrm{LLM}}(q \mid D,\, C,\, t^*,\, a^*)
    \label{eq:ask}
\end{equation}

The Ask step generates a clinical question $q$ conditioned on the dialogue history $C$, the reasoning chain $t^*$, and the selected strategy $a^*$. Three constraints govern generation. First, the question must follow naturally from the prior conversation without abrupt topic shifts. Second, the selected strategy must be applied implicitly --- the question should create the intended conversational conditions without naming the strategy or revealing the diagnostic intent. Third, the question must use patient-friendly natural language, free of clinical or diagnostic terminology that would be inappropriate in a real assessment context. Sampling temperature is set to $0.7$ to balance linguistic naturalness with response diversity.

\paragraph{Belief State and Feedback Loop} 
After each dialogue turn, the Trait Detector identifies SLD traits present in the patient response $(q, r)$ and updates the belief state maintained over each trait $f \in \mathcal{F}$. The belief state is represented as a Beta distribution $\mathrm{Beta}(\alpha_f, \beta_f)$, where $\alpha_f$ and $\beta_f$ accumulate positive and negative detection evidence respectively across turns. The posterior mean $\mu_f = \alpha_f / (\alpha_f + \beta_f)$ serves as the current probability estimate for trait $f$, and the entropy $H(\mathrm{Beta}(\alpha_f, \beta_f))$ quantifies residual uncertainty. Traits whose posterior mean exceeds a detection threshold $\tau$ are added to $\hat{\mathcal{F}}$ and fed back to the TPA Selector at the next turn, closing the proactive reasoning loop. This feedback mechanism ensures that the Think step's diagnostic gap analysis reflects the most current evidence state at every turn rather than a fixed initial assessment.

\subsection*{SLD Trait Detector}

Accurate identification of SLD traits from examiner--patient dialogue is a prerequisite for valid elicitation evaluation: if the Trait Detector fails to identify traits that are genuinely present in the patient response, coverage and F1 estimates will be artificially deflated regardless of how well the TPA Selector elicits them. We adopt the LLM-based trait detection framework developed in our prior work~\cite{hu2025exploiting}, which demonstrated that zero-shot LLM prompting outperforms supervised BERT-based classifiers on this task across sensitivity and positive predictive value.

At each dialogue turn, the Trait Detector receives the examiner question $q$ and patient response $r$, and identifies which of the ten SLD traits (F1--F10) are present in the patient's language. Detection is performed via structured zero-shot prompting of GPT-5.4-nano, conditioned on the full F1--F10 trait definitions (Tab.~\ref{tab:traits}) as domain knowledge. The prompt follows the two-component structure from \citet{hu2025exploiting}: the examiner--patient dialogue provides conversational context, and the trait definitions provide the diagnostic knowledge required to distinguish clinically relevant language patterns from normal conversational variation. The detector outputs a binary label for each trait $f \in \mathcal{F}$, indicating presence or absence in the patient response.

The output of the Trait Detector serves two functions within the TPA framework. First, it updates the confirmed trait set $\hat{\mathcal{F}}$, which is fed back to the TPA Selector at the next turn to inform the Think step's diagnostic gap analysis. Second, it provides the ground-truth-relative detection labels used to compute episode-level evaluation metrics --- Coverage, F1, and AUCC. True positives are traits detected that are in the ground-truth annotation $\mathcal{F}^*$; false positives are detected traits not in $\mathcal{F}^*$; false negatives are ground-truth traits not detected within the episode.

%%=============================================================
\section*{Discussion}
%%=============================================================

This study demonstrates that proactive diagnostic reasoning, implemented through the explicit Think-Plan-Ask cycle, substantially improves SLD trait elicitation in automated ADOS-2 language assessment. The central finding is that the performance gap between TPA and competitive dialogue planning baselines is not primarily attributable to architectural complexity or computational resources: several baselines employ more sophisticated search procedures than TPA, yet consistently underperform. Rather, the gains derive from the quality and transparency of the reasoning that precedes strategy selection. The following analyses examine the mechanisms underlying these gains across three dimensions: strategy reallocation, adaptive phase structure, and the relationship between per-turn elicitation rate and diagnostic completeness.

\textbf{Strategy reallocation, not per-strategy superiority,
drives coverage gains.}
The strategy-level analysis reveals a key mechanistic insight: TPA's advantage over real clinical dialogue does not stem from achieving higher elicitation rates within any individual strategy. Rather, the gains arise from TPA's systematic reallocation of turns away from a single dominant strategy toward strategies that target previously unelicited traits. Real clinical dialogue allocates $90\%$ of turns to Open-ended questioning regardless of which traits remain unconfirmed, resulting in repeated elicitation of already-surfaced traits and a coverage plateau after turn 10. TPA, guided by the Think step's explicit gap analysis, concentrates only $4\%$ of turns on Open-ended questioning and redistributes the remainder toward Correction-inducing ($43\%$), Hypothetical ($19\%$), and Multi-step ($14\%$) strategies --- each targeting traits that Open-ended questioning fails to surface reliably. Crucially, this reallocation is selective rather than mechanical: as shown in Fig. \ref{fig:strategy_comparison}, TPA deploys Open-ended questions at turns where novel trait discovery is most probable, rather than uniformly throughout the episode, which accounts for the sustained early-turn coverage growth visible in Fig. \ref{fig:coverage_curve}.

\textbf{Strategy selection adapts systematically across
the episode.}
The phase-level analysis reveals that TPA's strategy reallocation is not static but evolves in a characteristic three-phase pattern. In early turns (T1--5), TPA distributes effort broadly across multiple strategies --- Multi-step ($26\%$), Hypothetical ($16\%$), and Open-ended ($11\%$) --- reflecting an exploratory phase in which the Think step targets multiple unconfirmed traits simultaneously. From the mid phase (T6--12) onward, Correction-inducing questions rise sharply to $46\%$ and continue to dominate in late turns ($>50\%$), as the diagnostic gap narrows and the Think step redirects attention toward the most persistently unconfirmed traits, primarily echoic repetition (F1), which is rare but assigned high belief uncertainty when unobserved. This phase structure emerges directly from the Think step's per-turn belief state updates rather than from a fixed protocol, distinguishing TPA's adaptive allocation from the static Open-ended concentration observed in real clinical dialogue (Fig.  \ref{fig:strategy_phase}).

\textbf{High per-turn elicitation rates do not guarantee
high diagnostic coverage.}
A seemingly paradoxical finding emerges from the strategy
comparison: real clinical dialogue achieves comparable or
higher raw per-turn elicitation rates for several strategies,
yet achieves substantially lower overall trait coverage
(Tab. \ref{tab:tpa_vs_real}). This apparent contradiction is
resolved by distinguishing between trait elicitation and
novel trait elicitation. The per-turn elicitation
rate measures whether any SLD trait was detected in a
patient response, regardless of whether that trait had
already been confirmed in a prior turn. Real clinical
dialogue's elicitation rates therefore largely reflect
repeated detection of already-confirmed traits --- primarily
F2 (Unconventional Content) and F6 (Superfluous Phrase
Attachment), which are reliably surfaced by Open-ended
strategies from the earliest turns. Once these common traits
are confirmed, continued deployment of the same strategy
yields no additional coverage gain, and the coverage curve
flattens accordingly after turn 10. TPA, by contrast,
redirects turns toward Correction-inducing and Hypothetical
strategies whose absolute elicitation rates are lower
($0.04$--$0.08$) but whose marginal novelty is higher:
each deployment is more likely to surface a trait not yet
confirmed --- F1 (Echoic Repetition) and F3 (Pronoun
Displacement) --- than to re-detect one already known.
This distinction has a direct implication for how elicitation
efficiency should be measured: per-turn elicitation rate,
taken in isolation, is a misleading proxy for diagnostic
completeness. AUCC, which integrates coverage gain across
all turns and rewards systems that continuously surface
new traits rather than repeatedly confirming old ones, is
a more appropriate primary metric for evaluating
elicitation-focused assessment systems.

\textbf{SLD elicitation efficiency reflects task scope, not clinical expertise.}
The Coverage Gain Rate analysis reveals a striking divergence between which strategies are most effective for SLD trait elicitation and which strategies real clinicians actually deploy. Open-ended questioning achieves the highest Coverage Gain Rate ($0.22$) and Multi-step guidance ranks second ($0.15$), yet real clinical dialogue allocates $90\%$ of turns to Open-ended questioning alone and fewer than $2\%$ to Multi-step. This concentration on a single strategy --- however effective per turn --- cannot sustain coverage growth once common traits are confirmed, because repeated application of the same strategy surfaces the same traits. This divergence reflects the broader scope of real ADOS-2 assessment rather than a deficit in clinical expertise: full interviews must simultaneously establish rapport, complete standardised activities, and observe non-verbal behaviour --- purposes that consume turns without directly advancing SLD trait elicitation but are clinically indispensable. TPA, designed exclusively for elicitation within a constrained window, operates under no such constraint and can direct every turn toward closing the diagnostic gap. These conditions are not comparable, and the coverage difference should be interpreted accordingly.

\textbf{Explicit reasoning quality is the performance ceiling.}
The LLM sensitivity analysis reveals a steep monotonic relationship between backbone model capability and elicitation performance, spanning 34 percentage points in Coverage across the three models tested. This scaling behaviour reflects the clinical reasoning demands of the Think step: a model that cannot accurately identify which traits are likely present but unelicited, or reason about which conversational conditions would surface them, produces reasoning chains that misdirect the Plan step regardless of how well the Ask step executes the resulting question. The quality of explicit diagnostic reasoning is therefore the hard performance ceiling of the TPA framework --- a ceiling that rises as the capability of the underlying LLM improves, suggesting that TPA will benefit directly from continued advances in foundation model capability.

\textbf{Clinical implications.}
The $25\%$ relative improvement in trait coverage over real clinical dialogue, achieved within the same 20-turn budget, suggests that proactive strategy selection could substantially compress the diagnostic window for SLD trait elicitation in automated assessment applications. The interpretability of the Think step --- which produces an explicit, human-readable reasoning chain at every turn --- provides a degree of transparency that black-box planning approaches cannot offer, and is a prerequisite for clinical deployment where assessors need to understand and verify the reasoning behind automated recommendations. It is important to emphasise that TPA addresses a specific and well-defined sub-problem within autism assessment: the systematic elicitation of SLD traits within a constrained dialogue window. Its appropriate role is as a scalable supplementary tool for screening, training, or remote assessment applications where expert clinician time is unavailable, not as a replacement for trained ADOS-2 assessors whose role extends far beyond SLD trait elicitation.

\textbf{Limitations and future directions.}
Several limitations of the present study warrant acknowledgement. First, evaluation is conducted within a simulated environment: the Patient Agent, while validated for trait frequency alignment (AUC $= 0.750$) and semantic coherence (cosine similarity $= 0.400$), cannot fully replicate the complexity and variability of real patient language. Validation against real patient interactions is a necessary next step before clinical deployment. Second, the evaluation dataset comprises 35 verbally fluent adolescents and adults with confirmed ASD diagnoses; performance on populations with different linguistic profiles, co-occurring conditions, or lower expressive language levels remains to be established. Third, the strategy-level analysis reveals that five of the six strategies in real clinical dialogue --- Hypothetical, Correction, Emotion, Perspective, and Multi-step --- are deployed in fewer than $10\%$ of turns each, yielding insufficient sample sizes for reliable per-strategy elicitation rate estimates; the real-versus-simulated strategy comparison should therefore be interpreted with caution for these conditions. Fourth, the current framework operates on text transcripts and does not incorporate the non-verbal and prosodic signals that trained clinicians use to contextualise SLD trait judgements; integration of multimodal signals represents a promising direction for future work~\cite{ruan2023mmys,ruan2024can,zhang2022discriminative,yu2024video,hu2025speech}.

Future work will prioritise three directions: prospective
validation with real patients in controlled assessment
settings; extension to other structured psychiatric
interviews where specific linguistic features must be
actively elicited; and investigation of whether the
proactive reasoning approach generalises to longer episodes,
higher-trait-density patients, and non-English language
contexts.

\section*{Conclusion}

This work reframes automated autism language assessment as an elicitation problem rather than a detection problem. The central demonstration is that the bottleneck is not identifying traits once they appear in transcript, current LLM-based detectors do this reliably, but creating the conversational conditions under which latent traits emerge at all. TPA addresses this through an explicit Think-Plan-Ask reasoning cycle that produces human-readable justifications for every strategy decision, a property that black-box planning approaches cannot offer and that is a prerequisite for clinical trust.

The results carry a broader methodological implication: in any structured assessment where diagnostically relevant features must be actively elicited rather than passively observed, the quality of reasoning about what is still missing and why matters more than the sophistication of the search procedure used to act on that reasoning. This principle is not specific to autism or to SLD traits, it generalises to any clinical interview in which the evidence must be created by the interaction itself. We anticipate that future work validating TPA with real patients, extending it to other psychiatric interviews, and integrating multimodal signals will build on this foundation toward scalable, interpretable AI-assisted assessment across neurodevelopmental and psychiatric conditions.

\section*{Acknowledgements}

\backmatter

%\bmhead{Supplementary information}

%If your article has accompanying supplementary file/s please state so here. 

%Authors reporting data from electrophoretic gels and blots should supply the full unprocessed scans for key as part of their Supplementary information. This may be requested by the editorial team/s if it is missing.

%Please refer to Journal-level guidance for any specific requirements.

%\bmhead{Acknowledgements}

%Acknowledgements are not compulsory. Where included they should be brief. Grant or contribution numbers may be acknowledged.

%Please refer to Journal-level guidance for any specific requirements.

\section*{Declarations}

%Some journals require declarations to be submitted in a standardised format. Please check the Instructions for Authors of the journal to which you are submitting to see if you need to complete this section. If yes, your manuscript must contain the following sections under the heading `Declarations':

\begin{itemize}
\item Acknowledgments

%This research was supported by the NSF (HCC-2401748) and NIH (R01MH129426). 
This material is based upon work supported by the National Science Foundation under Grant No. HCC-2401748 and the National Institute of Health under Grant No. R01MH129426. The funders had no role in study design, data collection and analysis, decision to publish, or preparation of the manuscript.

\item Author contribution
C.H., M.Y., B.L., S.W., X.L. designed research. C.H. performed experiments. C.H., M.Y., W.L., and L.K.P. analyzed data. C.H., B.L., S.W., and X.L. wrote the paper. All authors have read and approved the manuscript.

%\item Competing interests 

%The authors declare no competing interests. 
\end{itemize}

\section*{Data and Code Availability Statement} 

The code accompanying this research is publicly available on GitHub: {\url{https://github.com/cbhu523/Multi-Agent-System-for-Autism}}\\

\section*{Conflict of Interest Statement}
The authors declare no competing interests.

\noindent

\bibliography{sn-bibliography}% common bib file

\newpage

\end{document}